\documentclass[twocolumn]{svjour3}          

\usepackage{fix-cm}
\usepackage[T1]{fontenc}
\usepackage{amsmath,amssymb,amsfonts}
\usepackage{algorithmic}
\usepackage{graphicx}
\usepackage{textcomp}
\usepackage{diagbox}
\usepackage{float}
\usepackage{url}
\usepackage{amsfonts}
\usepackage{rotating}
\usepackage{multirow}
\usepackage{hhline}
\usepackage{adjustbox}
\usepackage{fancyhdr}
\usepackage{rotating}
\usepackage{hyperref}
\usepackage{bm}
\usepackage{makecell}
\usepackage{mleftright}
\usepackage{comment}
\usepackage{algorithm2e}
\usepackage{marvosym}
\RestyleAlgo{ruled}
\usepackage{cite}

\DeclareMathOperator*{\argmin}{arg\,min}  

\begin{document}

\title{High Speed Neuromorphic Vision-Based Inspection of Countersinks in Automated Manufacturing Processes}

\titlerunning{High Speed Neuromorphic Vision-Based Inspection}

\author{Mohammed Salah$^{1,2,*,\dagger}$ \and Abdulla Ayyad$^{1,\dagger}$ \and Mohammed Ramadan$^{1}$ \and Yusra Abdulrahman$^{1,3}$ \and  Dewald Swart$^{4}$ \and Abdelqader Abusafieh$^{4}$ \and Lakmal Seneviratne$^{2}$ \and Yahya Zweiri$^{1,3}$}

\authorrunning{Salah et al.} 
\institute{This work was supported by by the Advanced Research and Innovation Center (ARIC), which is jointly funded by STRATA Manufacturing PJSC (a Mubadala company), Khalifa University of Science and Technology in part by Khalifa University Center for Autonomous Robotic Systems under Award RC1-2018-KUCARS, and Sundooq Al Watan under Grant SWARD-S22-015. \\
             $^{1}$ 
              Advanced Research and Innovation Center, Khalifa University, Abu Dhabi, UAE \\
             $^{2}$
              Khalifa University Center for Autonomous Robotic Systems (KUCARS), Khalifa University of Science and Technology, Abu Dhabi, United Arab Emirates \\
              $^{3}$
              Department of Aerospace Engineering, Khalifa University, Abu Dhabi, United Arab Emirates \\
              $^{4}$
              Research and Development, Strata Manufacturing PJSC, Al Ain, UAE. \\
            $\dagger$ indicates equal contribution to this work. \\ $^{*}$ is the corresponding author. \\ \email{mohammed.salah@ku.ac.ae}}

\date{Received: date / Accepted: date}

\maketitle

\begin{abstract} 
Countersink inspection is crucial in various automated assembly lines, especially in the aerospace and automotive sectors. Advancements in machine vision introduced automated robotic inspection of countersinks using laser scanners and monocular cameras. Nevertheless, the aforementioned sensing pipelines require the robot to pause on each hole for inspection due to high latency and measurement uncertainties with motion, leading to prolonged execution times of the inspection task. The neuromorphic vision sensor, on the other hand, has the potential to expedite the countersink inspection process, but the unorthodox output of the neuromorphic technology prohibits utilizing traditional image processing techniques. Therefore, novel event-based perception algorithms need to be introduced. We propose a countersink detection approach on the basis of event-based motion compensation and the mean-shift clustering principle. In addition, our framework presents a robust event-based circle detection algorithm to precisely estimate the depth of the countersink specimens. The proposed approach expedites the inspection process by a factor of 10$\times$ compared to conventional countersink inspection methods. The work in this paper was validated for over 50 trials on three countersink workpiece variants. The experimental results show that our method provides a precision of 0.025 mm for countersink depth inspection despite the low resolution of commercially available neuromorphic cameras. 
\href{https://www.dropbox.com/s/pateqqwh4d605t3/final_video_new.mp4?dl=0}{Video Link}

\end{abstract}


\section{INTRODUCTION}

Automation of manufacturing processes has been transforming the industrial landscape for over a century, beginning with Ford's introduction of the assembly line. While automated manufacturing was initially aimed at producing large quantities, low-value, and standardized items \cite{karim2013challenges}, it is now being applied to producing high-value, specialized items. This shift is driven by the need to achieve and sustain competitiveness in an operating environment with a growing number of competitors and declining highly specialized labor \cite{jayasekara2022level}. In addition, automated manufacturing is becoming increasingly feasible due to the combination of low-cost computational power and the declining costs of sensors and actuators. This allows for more specialized applications to be realized across various industries, including aerospace and automotive assembly lines \cite{jayasekara2022level}. These technologies are making it possible to create flexible, efficient, and lean automated manufacturing processes.

Automated industrial inspection is one of the essential operations in automated assembly lines, with computer vision being the driving force of the aforementioned process \cite{cim_survey, inspection_jim_machinevision}. Such fundamental operation suffices the objectives of safety and quality control, allowing industries to identify the potential faults of products and conduct corrective action accordingly. Therefore, industrial inspections demand precise scanning for complete, digital representation of mechanical parts. 3D scanning is widely used for geometric reconstruction, and automated aircraft inspections \cite{yasuda2022aircraft, da20233d, shahid_hybrid, jim_auto_inspection}. Different types of scanners with varied capabilities in terms of accuracy and portability have been developed for visual inspection in manufacturing. Kruglova et al. \cite{kruglova2015robotic} developed a drone embedded 3D scanner for aircraft wing inspection. However, they reported challenges with achieving the required levels of precision. Sa et al. \cite{sa2018design} developed visual measurement technology for counterbores based on laser sensors. Vision inspection is also being used in inspecting drilling holes in aircraft manufacturing with acceptable levels of accuracy \cite{3d_csk_measurement}. Other applications of computer vision in an inspection include: trajectory generation \cite{phan2018path} and robot-attached 3D scanners for inspection and repair, respectively \cite{burghardt2017robot, cladding}. 

Countersink inspection is a critical aspect of drilling and requires reliable and fast inspection methods for quality control. As such, it is important to integrate the inspection process into an automation system to achieve high production rates, enhanced quality control accuracy, and ensure traceability in real time \cite{integrated_hole}. According to Luker et al. \cite{luker2020process}, countersink depth, and the associated flushness of the fastener head is one of the challenging aspects to inspect and automate using computer vision approaches. The requirement for precise tolerance levels is restricted by the limitations of available laser scanning systems in terms of cost, and the lack of controlled industrial operating environments \cite{luker2020process}. Furthermore, considering the wide adoption of countersinks in flush rivets of aircraft paneling, inspection for quality control significantly impacts the performance of riveted joints \cite{yu2019vision}, imposing a surging demand for developing a reliable countersink inspection system.

Countersink inspection is mainly undertaken with the aid of two methods: non-contact and contact. The contact approach, basically the same as mechanical probing, is mainly used to determine the interior diameter of holes \cite{yu2019vision}. The main challenge with the contact method is high susceptibility to loss of accuracy resulting from debris and cutting lube buildup \cite{nsengiyumva2021advances, yu2019vision}. Non-contact methods include laser sensors and machine vision, which are increasingly adopted with current industrial automation trends. Haldimann \cite{3d_csk_measurement} performed countersink inspection using digital light processing (DLP) projector and a monocular camera achieving a precision of 0.03 mm. In addition, Luker \& Stansbudy \cite{luker2020process} proposed using high-accuracy laser scanners to inspect countersinks, dropping the precision up to 6 $\mu m$. While these works provide the requirements of the current industrial requirements, they remain costly active methods and require extensive calibration and instrumentation. Therefore, Yu et al. \cite{yu2019vision} utilized a high-resolution monocular camera for countersink inspection, while abiding by the industry's requirements and attaining a precision of 0.02 mm. Nevertheless, such precision is conditional, given the imaging sensor is normal to the countersink specimens. To circumvent this limitation, Wang and Wang \cite{wang2016handheld} utilized spectral domain optical coherence tomography (SD-OCT) for non-contact countersink inspection to deliver accurate countersink measurements without requiring precise scan center alignment.

Despite the growing adoption of computer vision for automated countersink inspection, the execution time of the inspection task was never taken into consideration. In all of the aforementioned methods, sensor motion during inspection is prohibited as it introduces measurement errors and degrades precision. This restriction results from the fact that laser scanners are associated with high latency and conventional cameras suffer from motion blur, leading to prolonged execution time of the inspection task. Neuromorphic vision sensors (NVS) address the limitations of the aforementioned sensing pipelines delineated by their low latency ($1 \mu s$), high temporal resolution ($1 ms$), and high dynamic range ($120$ dB), granting them robustness against motion blur and illumination variation \cite{vision_transformer, davis}. Unlike synchronous conventional cameras, the NVS generates asynchronous events as a response to variations of light intensity \cite{event_survey}. This asynchronous nature triggered a paradigm shift in the computer vision community, as neuromorphic cameras unveiled their impact in space robotics \cite{nvbm_tim, scramuzza_space, space_debris}, robotic manufacturing \cite{ayyad_drilling}, robotic grasping \cite{flicker_grasp, grasp_wong}, autonomous navigation \cite{event_odometry_1, event_odometry_2}, tactile sensing \cite{yahya_tactile, yahya_tactile_2}, and visual servoing \cite{hay2021unified}.

\subsection{Contributions}

With the unprecedented capabilities of neuromorphic cameras in hand, this paper investigates the feasibility of high-speed neuromorphic vision-based inspection of countersinks. In the proposed framework, the NVS asynchronous nature is leveraged, where the sensor rather sweeps the countersinks on the mechanical part instead of pausing on each hole, reducing the inspection time to a few seconds. However, the NVS is an emerging technology and requires event-based perception algorithms for neuromorphic perception. Accordingly, this article comprises the following contributions to neuromorphic inspection of countersinks:

\begin{enumerate}
    \item A novel event-based countersink detection algorithm is proposed based on motion compensation and the mean-shift clustering principle that overcomes the speed limitations of conventional countersink inspection methods.
    \item A robust event-based circle detector is devised on the basis of Huber loss minimization to cope with the unconventional nature of the NVS and to provide the required inspection precision.
    \item To validate the proposed inspection framework, we performed over 50 trials of an NVS sweeping three countersink workpieces with different hole sizes at speeds ranging from 0.05 to 0.5 m/s. The results show the inspection process is accelerated by a factor of 10$\times$ compared to state-of-the-art inspection methods while providing a precision of 0.025 mm with a low resolution NVS.
\end{enumerate}

\subsection{Structure of The Article}
The rest of the article is organized as follows. Section \ref{sec:preliminaries} provides an overview of the neuromorphic vision working principle and an overview of the inspection setup. Section \ref{sec:methodology} discusses the theory behind the event-based motion compensation and event-based circle detection for countersink depth estimation. Experimental validation of the proposed inspection pipeline is reported in section \ref{sec:exps}, and section \ref{sec:conc} presents conclusions and future works.

\begin{figure*}[!ht]
    \centering
    \includegraphics[keepaspectratio=true,scale=0.45, width=\linewidth]{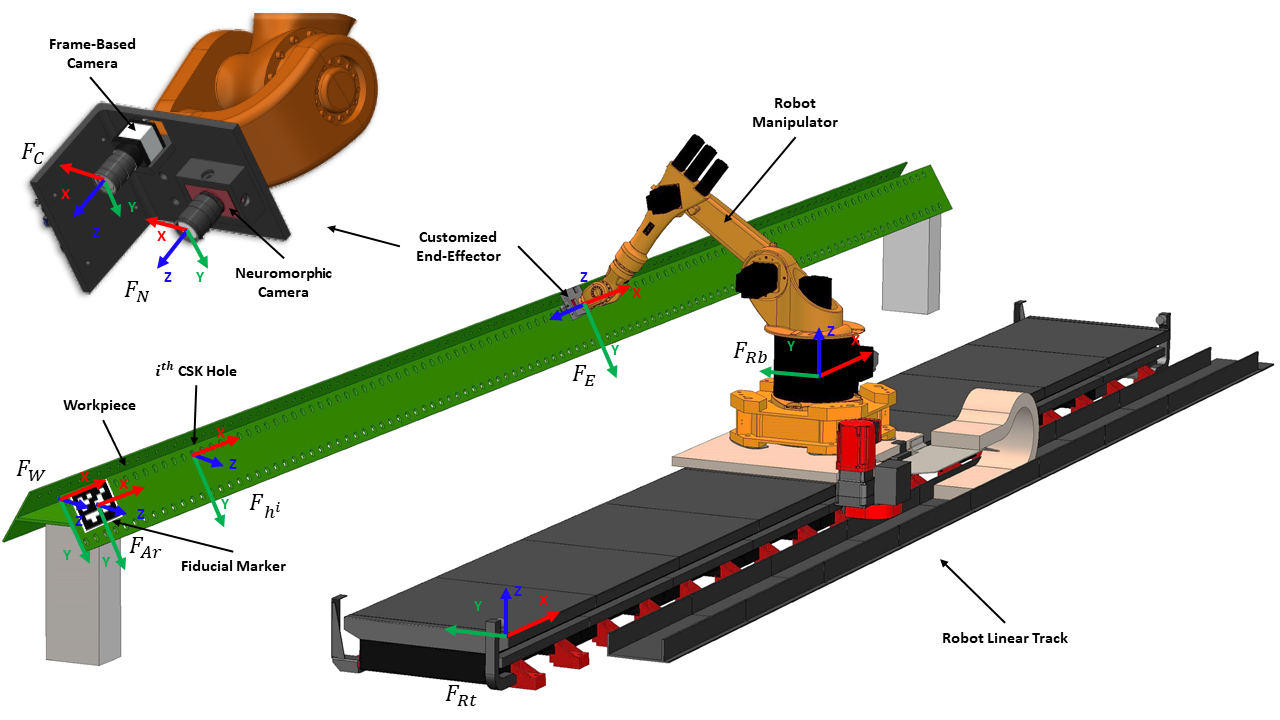}
    \caption{The robotic inspection setup illustrates the different components and defines the coordinate frames used for navigation and control.}
    \label{fig:Robotic_Setup}
\end{figure*}

\section{Preliminaries} \label{sec:preliminaries}
\subsection{Robotic Inspection Setup} \label{sec:overview}
The overall setup of the neuromorphic vision-based inspection system can be seen in Figure \ref{fig:Robotic_Setup}. The presented system consists of an industrial robotic manipulator, and a customized end-effector holding a frame-based camera and a neuromorphic camera.

For the purpose of robot navigation and control, we have defined the following frames of reference:
\begin{itemize}
    \item \(\mathcal{F}_{Rb}\): The robot base coordinate frame.
    \item \(\mathcal{F}_{Rt}\): The robot linear track coordinate frame.
    \item \(\mathcal{F}_E\): The robot end-effector coordinate frame.
    \item \(\mathcal{F}_C\): The frame-based camera coordinate frame.
    \item \(\mathcal{F}_N\): The neuromorphic camera coordinate frame.
    \item \(\mathcal{F}_W\): The workpiece coordinate frame.
    \item \(\mathcal{F}_{Ar}\): The fiducial / ArUco coordinate frame.
    \item \(\mathcal{F}_{h^i}\): The coordinate frame of the $i^{th}$ reference hole.
\end{itemize}

The rotation matrix mapping a source frame \(\mathcal{F}_S\) to a target frame \(\mathcal{F}_T\) can be defined as \(_{T}R_{S} \in \mathbb{R}^{3\times3}\). The position vector that gives the relative position of point b to point a described in \(\mathcal{F}_T\) can be defined as \(^{T} _{a}\vec{P}_{b} \in \mathbb{R}^3\). Then, we can define the affine transformation  \(_{T}T_{S} \in \mathbb{R}^{4\times4}\) as follows:

\begin{equation}
    \label{eq_affine_transformation}
    _{T}T_{S} = \begin{bmatrix}
    _{T}R_{S} & ^{T}_{T}\vec{P}_{s} \\
    \mathbf{0}^T & 1
    \end{bmatrix}
\end{equation}

Using the robot's forward kinematics relations, the affine transformation matrix $_{Rt}T_{E}$ can be identified and considered known for the remainder of this paper. The robot's forward kinematics can be denoted as follows:

\begin{equation}
    \label{eq_forward_kinematric}
    _{Rt}T_{E} = g(\theta),  \hspace{5mm} \theta \in \mathbb{C}
\end{equation}
where  $\theta$ is the monitored  angles of the robotic manipulator and linear track, $g(\theta)$ is a nonlinear function representing the robot's kinematics, and $\mathbb{C}$ is the robot's configuration space. Additionally, the constants $_{E}T_{C}$ and $_{E}T_{N}$ can be computed using the geometrical calibration method explained in \cite{ayyad2023neuromorphic}. Hence, $_{B}T_{E}$ and the calibrated transformations can be utilized to solve for $_{B}T_{C}$ and $_{B}T_{N}$. Finally, we define the neuromorphic camera's twist vector $\vec{V}_{N} = [v_x, v_y, v_z, \omega_x, \omega_y, \omega_z]^T$ that represents the neuromorphic camera's velocity in its own frame \(\mathcal{F}_N\), where $[v_x, v_y, v_z]$ and $[\omega_x, \omega_y, \omega_z]$ are the linear and angular velocities respectively. The velocity components are computed from the forward kinematics of the robot manipulator as follows:
\begin{equation}
    \vec{\mathcal{V}}_N = J(\theta)\dot{\theta}
    \label{eq_vel_forward_kinematics}
\end{equation}
\begin{equation}
J(\theta) \triangleq \dfrac{\partial g_{N}}{\partial \theta} \in \mathbb{R}^{6 \times N_j}
 \label{jacobian}
\end{equation}
where \(J(\theta)\) is the Jacobian matrix of the camera's forward kinematics function \(g_N\), and \(N_j\) is the number of robot joints.

As the major reference frames of the robotic setup are defined, the inspection routine is performed, with the neuromorphic camera sweeping the workpiece at a constant high speed. Prior to initiating the inspection process, the robot must first localize the workpiece in 6-DoF and align the neuromorphic camera accordingly. The robot starts with a priori estimate of the workpiece's pose \(_{Rt}\hat{T}_{W}\), and the pose of the fiducial marker \(_{Rt}\hat{T}_{Ar}\). The robot then places the frame-based camera with a specific standoff away from the fiducial marker and utilizes the ArUco pose estimation method of the OpenCV library \cite{openCV} to refine \(_{Rt}\hat{T}_{Ar}\), similar to the approach used in \cite{halawani_visuo}. Subsequently, the pose of each $i^{th}$ hole relative to the fiducial marker is known from the CAD model of the workpiece and is used to estimate the hole's pose relative to the robot's base frame:
\begin{equation}
    _{Rt}\hat{T}_{h^i} =  \hspace{1pt} _{Rt}\hat{T}_{Ar} \hspace{1pt} _{Ar}T_{h^i} , \hspace{0.5cm} i \in [1, N_h]
    \label{eq_hole_initial}
\end{equation}

Once the pose of each hole is estimated, the robot places the neuromorphic camera normal to the first hole with a specific relative pose and initiates the sweeping process. The sweeping process is performed by moving the neuromorphic camera with a constant velocity \(\vec{V}_N^*\) across all the holes that lie in a straight line and have identical orientation. Given the position of the first hole \(^{N} _{N}\vec{P}_{h^1}\) and last hole \(^{N} _{N}\vec{P}_{h^{N_h}}\) described in \(\mathcal{F}_N\), the linear components of \(\vec{V}_N^*\) are computed as follows:

\begin{equation}
    \label{eq_command_vel}
    \begin{bmatrix}
    v^*_x \\
    v^*_y \\
    v^*_z \\
    \end{bmatrix} = v^* \hspace{0.2cm} \frac{^{N} _{N}\vec{P}_{h^{N_h}} - ^{N} _{N}\vec{P}_{h^{1}}} {|| ^{N} _{N}\vec{P}_{h^{N_h}} - ^{N} _{N}\vec{P}_{h^{1}} ||}
\end{equation}
where $v^*$ is the desired magnitude of the sweeping speed. The rotational components of the sweeping motion are set to zero $[\omega^*_x, \omega^*_y, \omega^*_z]^T = \vec{0}$. The commanded camera velocity \(\vec{V}_N^*\) is then transformed to command joint velocities $\theta^*$ by inverting the expression in \eqref{eq_vel_forward_kinematics}, which are tracked using low-level PID controllers.



\subsection{Neuromorphic Vision Sensor} \label{sec:nvs}
\label{subsection:neuromorphic_sensor}
The NVS is a bio-inspired technology that comprises an array of photodiode pixels triggered by light intensity variations. Upon activation, the pixels generate asynchronous events, where they are designated with spatial and temporal stamps as

\begin{equation}
    e_k\doteq(\vec{x}_k,t_k,p_k),
\end{equation}

\noindent where $(x, y)_{k} \in W \times H$ is the spatial stamp defining the pixel coordinates, \(W\in\{1,...,w\}\) and \(H\in\{1,...,h\}\) represent the sensor resolution, \(t\) is the timestamp, and \(p\in\{-1,+1\}\) is the polarity corresponding to a decrease and increase in light intensity, respectively.

While the NVS output is unorthodox in contrast with traditional imaging sensors, it is fundamentally important to point out that the NVS utilizes identical optics to conventional cameras. Thus, the neuromorphic sensor follows the standard perspective projection model expressed by

\begin{equation}
        s\begin{bmatrix}
        u \\
        v \\
        1
    \end{bmatrix} = \underbrace{
    \begin{bmatrix}
f{x} & \gamma & u_{0}\\
0 & f_{y} & v_{0}\\
0 & 0 & 1
\end{bmatrix}
    }_{K} \underbrace{
    \begin{bmatrix}
r_{11} & r_{12} & r_{13} & t_{1}\\
r_{21} & r_{22} & r_{23} & t_{2}\\
r_{31} & r_{32} & r_{33} & t_{3}
\end{bmatrix}}_{R|\mathbf{t}} \begin{bmatrix}
x \\
y \\
z \\
1
\end{bmatrix},
\label{eq:projection}
\end{equation}

\noindent where $u$ and $v$ are the image coordinates of a mapped point in 3D space with world coordinates $x$, $y$, $z$, whereas $K$ is the intrinsic matrix comprising of the focal lengths $f_{x}$ and $f_{y}$, skew $\gamma$, and principal point $[u_{0}, v_{0}]$. On the other hand, $R|t$ is the extrinsic parameters representing the transformation from world coordinates to the sensor frame.

\section{Methodology} \label{sec:methodology}
This section presents the event-based inspection pipeline of the countersink holes. As the NVS performs a quick sweep of the countersinks workpiece, the unconventional nature of the neuromorphic sensor permits utilizing conventional image processing techniques for inspection. Therefore, section \ref{sec:detection} presents an event-based method for identifying the countersinks on the basis of motion compensation and mean-shift clustering. Finally, we propose a robust circle detection algorithm to estimate the depths of the countersink specimens with high precision, as described in section \ref{sec:depth}.

\subsection{Event-Based Detection of Countersinks} \label{sec:detection}
As the neuromorphic vision sensor sweeps the workpiece, let the $N_{e}$ events in $e_{k}$ $\in$ $\mathbf{E}_{i}$ be the generated events when the sensor is normal to a designated countersink set. Since dealing with $\mathbf{E}_{i}$ in the space-time domain introduces undesired computational complexities, the events need to be transformed to a 2D space for lightweight feature extraction of the observed holes. A naive approach is to accumulate the events to 2D images directly from their spatial stamps $\vec{x}_{k}$. Nevertheless, large accumulation times are correlated with motion blur, while short accumulation periods lead to degraded features of the countersink specimens. To counteract these limitations, we rather adopt event-based motion compensation \cite{unifying_contrast, moseg} and leverage the asynchronous nature of neuromorphic events to create sharp images of \textit{warped} events (IWE) defined by

\begin{equation}
    H_{i}(\vec{x};\vec{\Dot{U}}) = \sum^{N_{e}}_{k}p_{k}\delta(\mathbf{x} - \vec{x}'_{k}),
\end{equation}

\noindent where
\begin{equation}
    \vec{x}'_{k} = \vec{x}_{k} - \vec{\Dot{U}} (t_{k} - t_{i}).
\end{equation}

\noindent Each pixel $\mathbf{x}$ in the IWE $H_{i}$ sums the polarities $p_{k}$ of the warped events $\vec{x}'_{k}$ to a reference time $t_{i}$ given a candidate optical flow $\vec{\Dot{U}} = [\Dot{u}, \Dot{v}]^{T}$, where $t_{i}$ is the initial timestamp of $\mathbf{E}_{i}$ and $\delta$ is the dirac delta function. As discussed in section \ref{sec:overview}, the inspection involves a fixed countersinks workpiece, while the NVS twist vector $\vec{V}_{N}$ and the workpiece depth $Z$ are known as a priori. Therefore, $\vec{\Dot{U}}$ can be computed using the image jacobian $\mathbf{J}$ \cite{handbook} as

\begin{figure}[t]
\center
\includegraphics[scale=0.3]{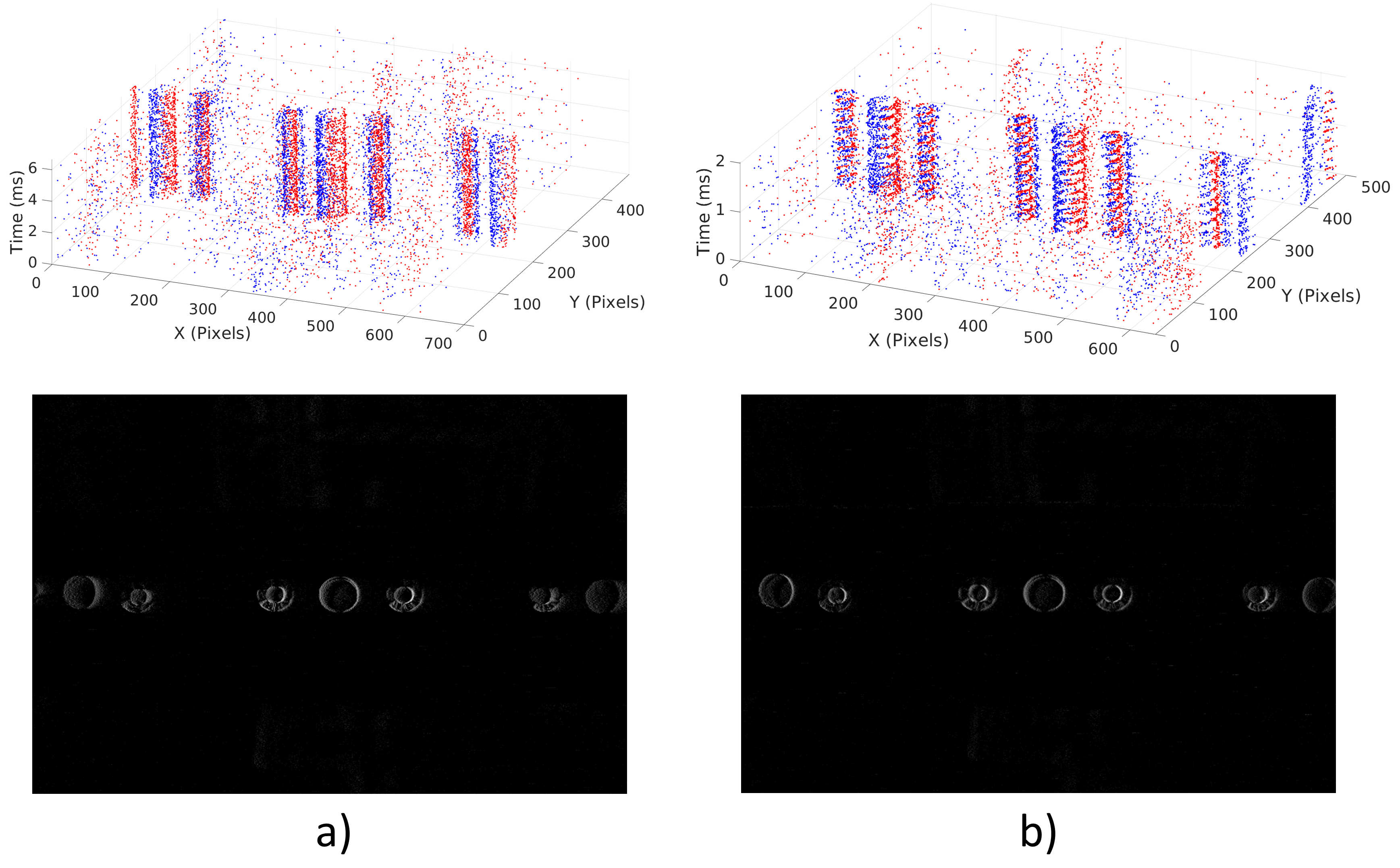}
\caption{a) The raw event stream and the constructed $H_{i}$ at 0.05 m/s, and b) presents the raw event stream and $H_{i}$ at 0.5 m/s. Both IWEs are textured regardless of the NVS velocity.}
\label{fig:iwe_events}
\end{figure}

\begin{equation}
    \vec{\Dot{U}} = \underbrace{\begin{bmatrix}
    \frac{-F}{Z} & 0 & u/Z & uv/F & -(F+u^2/F) & v \\
    0 & \frac{-F}{Z} & v/Z & F+v^2/F & -uv/F & -u
    \end{bmatrix}}_{\mathbf{J}} \vec{V}_{N}, \label{eq:jac}
\end{equation}

\normalsize
\noindent where \(F\) is the focal length and $[u,v]$ are the pixel coordinates. We constrain the NVS motion during the sweep to 2D linear motion reducing the expression in (\ref{eq:jac}) to $\Dot{u} = \frac{-F}{Z}v_{x}$ and $\Dot{v} = \frac{-F}{Z}v_{y}$. Having $\vec{\Dot{U}}$ expressed in closed form, $H_{i}$ is simply constructed at each timestep $i$. It is worth mentioning that in our approach, $\mathbf{E}_{i}$ comprises of at least $N_{e} = 10000$ events to create a textured $H_{i}$ regardless of its time step size.  Figure \ref{fig:iwe_events} shows the raw event stream along with the obtained $H_{i}$ at speeds of 0.05 and 0.5 m/s. Notice that the quality of the IWEs is maintained regardless of the sensor speed.

After constructing $H_{i}$, an opening morphological transformation \cite{morph} is applied to remove background noise, and the remaining activated pixels of warped events $e'$ are what we call the surface of active events (SAE). To identify the normal countersinks set present in $H_{i}$, $e'$ is fed to unsupervised mean-shift clustering algorithm \cite{mean_shift} with a tunable bandwidth $\beta$ and a flat kernel $\mathcal{K}(e')$ defined by

\begin{equation}
    \mathcal{K}(e') = \frac{1}{2} \begin{cases}
    1,& \text{if } -1 \le |G| \le 1\\
    0,              & \text{otherwise,}
\end{cases}
\end{equation}

\noindent and

\begin{equation}
    G = |\frac{e'-e_{k}}{\beta}|\text{ }, \text{where } i = 1, 2, \cdots, N_{e}.
\end{equation}

\begin{figure}[ht]
\center
\includegraphics[scale=0.25]{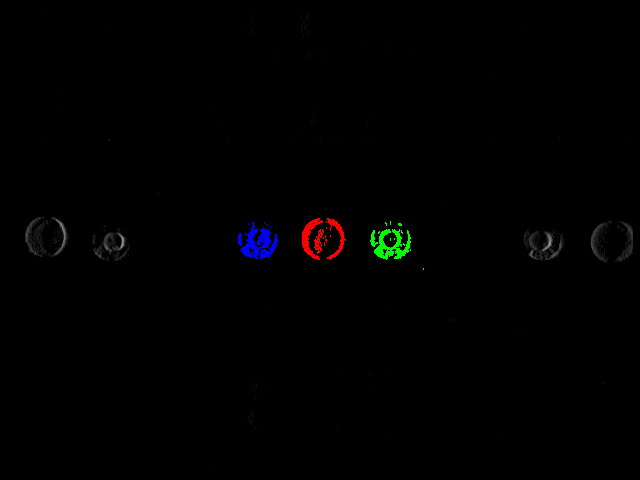}
\caption{Identified countersinks for nut-plate installation holes during the inspection using the mean-shift clustering principle. The left and right countersinks are highlighted in blue and green, respectively. The red points represent the pilot hole.}
\label{fig:csk_id}
\end{figure}

\noindent Given $\mathcal{K}(e')$, $l_{1} \cup l_{2} \cup \cdots \cup l_{L}$ mutually exclusive clusters are obtained by optimizing the clustering parameters, as illustrated in \cite{mean_shift}. Since multiple countersinks are observed at time step $i$, the normal countersinks to be inspected correspond to the cluster $\mathbf{L^{*}}$ closest to the principal point expressed by

\begin{equation}
    \mathbf{L^{*}} =  \argmin_{L}|| \vec{W}_{l} - \begin{bmatrix}
        u_{0} \\ 
        v_{0}
    \end{bmatrix} ||_{2},
\end{equation}

\noindent where $\vec{W}_{l}$ are the centroids of the clustered activated events $e'_{l}$. Figure \ref{fig:csk_id} shows the mean-shift clustering performance on a pilot hole and two nut-plate installation holes.

\subsection{Depth Estimation of Countersink Holes} \label{sec:depth}
After identifying the $l^{*}$ countersinks, their inner and outer radii need to be evaluated for depth estimation. Nevertheless, conventional circle detectors such as the circle Hough transform (CHT) and edge drawing-circles (ED-Circles) \cite{edge_drawing, ed_circles} are not applicable as an edge image is obtained with the absence of the gradient. Therefore, we devise a nonlinear robust least squares problem, as described in Algorithm \ref{alg:lsq} to jointly cluster $e'_{l}$ to either edge of the countersink and optimize the fitting parameters $\Gamma$, comprising of the circle inner $r$ and outer $R$ radii, and its image point center $h$ and $k$.

Given initial fitting parameters $\mathbf{\gamma}^{*} = [r, R, h, k]_{0}^{T}$, $e'_{l}$ is first transformed to polar coordinates $\vec{x}_{r} = || \vec{x'}_{l} - [h, k]_{0}^{T}||_{2}$, where $\vec{x'}_{l}$ are the spatial stamps of $e'_{l}$. Consequently, $\vec{x'}_{l}$ is clustered with hard data association by

\begin{equation}
    \vec{x}^{i}_{r} = \argmin_{r^{*}}|| \vec{x}_{r} - r ||_{2} \\
    \label{eq:radial_clust}
\end{equation}

\noindent where $\vec{x}^{i}_{r}$ are the activated events corresponding to the inner edge of the, while the rest of the events are otherwise. It is fundamentally important to highlight that the clustering in expression \ref{eq:radial_clust} is done at each optimization step. By jointly optimizing the clustering along with the fitting parameters, the expected likelihood of $e'_{l}$ probability density function is maximized, ensuring robust data association.

Finally, the residual vector $\vec{\xi}$ is formulated with the euclidean distance between the clustered $\vec{x}_{r}$ in \ref{eq:radial_clust} and $\gamma^{*}$ by

\begin{equation}
    \vec{\xi} = \begin{bmatrix} (\vec{X}^{i} - h_{0})^{2} + (\vec{Y}^{i} - k_{0})^{2} - r^{2} \\ (\vec{X}^{o} - h_{0})^{2} + (\vec{Y}^{o} - k_{0})^{2} - R^{2} \end{bmatrix},
\end{equation}

\noindent where $\vec{X}^{i}$ and $\vec{Y}^{i}$ are the spatial stamps of $\vec{x}^{i}_{r}$, while  $\vec{X}^{o}$ and $\vec{Y}^{o}$ are the spatial stamps of $\vec{x}^{o}_{r}$.

\begin{figure}[ht]
\center
\includegraphics[scale=0.4]{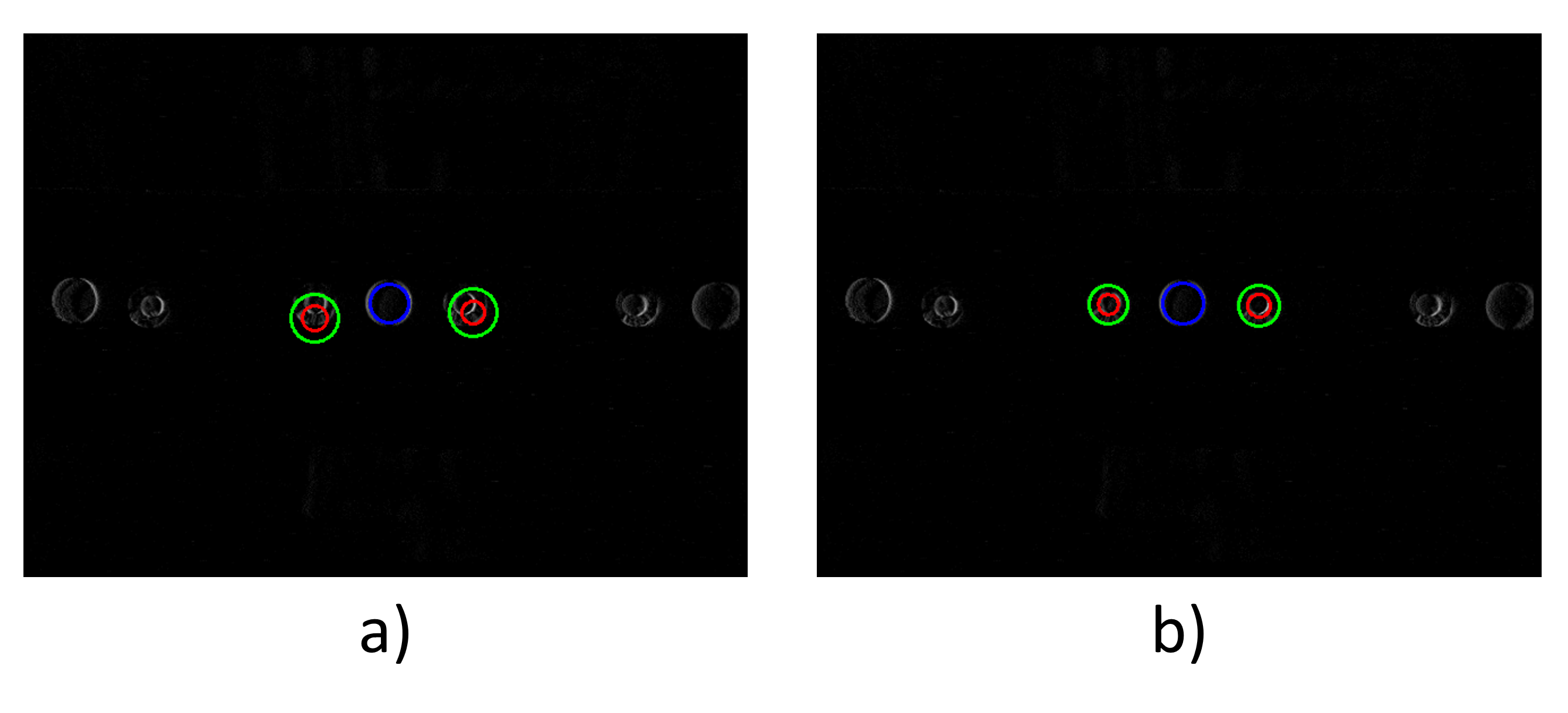}
\caption{a) Circle fitting using ordinary least squares and b) robust fitting by Algorithm \ref{alg:lsq}. The red and green circles represent the inner and outer edges of the. The blue circle is the pilot hole fitted using OLS.}
\label{fig:circle_fit}
\end{figure}

Utilizing ordinary least squares (OLS) to optimize $\Gamma$ can be affected by the presence of outliers. It is evident in Figures \ref{fig:iwe_events} and \ref{fig:csk_id} that the countersink circular edge is accompanied by Gaussian noise. While utilizing weighted least squares with a normally distributed weighting vector can alleviate this restriction, reflections also appear on the inner walls of the holes, degrading the aforementioned approach's performance. Instead of minimizing the sum of the squared residuals, a Huber loss function \cite{huber} is utilized to penalize the impact of outliers. The Huber loss $\alpha(\vec{\xi})$ is formulated as

\begin{equation}
    \alpha(\vec{\xi}) = \begin{cases}
    \vec{\xi}^{2},& \text{if } |\vec{\xi}| \le 1 \\
    2\vec{\xi} - 1,              & \text{otherwise,}
\end{cases}
\label{eq:huber}
\end{equation}

\noindent and is fed to the Levenberg–Marquardt (LM) \cite{lm} algorithm to optimize $\Gamma$ obtained as

\begin{equation}
    \Gamma = \begin{bmatrix}
    r \\
    R \\
    h \\
    k \\
    \end{bmatrix} = \argmin_{\Gamma^{*}}\sum_{l} \alpha(\vec{\xi}),
    \label{eq:cost}
\end{equation}

\noindent where the optimization is initialized with initial condition $\gamma^{*}$ parameterized by the mean of the inner $r_{0}$ and outer radii $R_{0}$, and the centroid of $\vec{x}_{l}$. Figure \ref{fig:circle_fit} shows the circle fitting using Huber loss and is compared to fitting using OLS on nut-plate installation holes. Notice that the fitting is robust against the noise and reflection outliers, unlike the linear loss of OLS.

\begin{algorithm}[t]
\caption{Depth Estimation of Countersink Holes}
\label{alg:lsq}
\SetKwInput{Input}{Inputs~}
\SetKwInOut{Output}{Output}
\Input{$e'_{l} \in l^{*}$}
\KwData{$Z$, $\phi$}
\Output{$\Gamma = [r, R, h, k], Z'$}
\SetKw{Initialize}{Initialize}
\Initialize{$\gamma^{*} = [r_{0}, R_{0}, h_{0}, k_{0}]^{T}$} \\
\For{$l=1$ \KwTo $l^{*}$}{
\While{\textup{Not Converged}}{
\textup{Transform }$\vec{x}_{l}$ to $\vec{x}_{r} = || \vec{x'}_{l} - [h, k]_{0}^{T}||_{2}$ \\
\textup{Cluster} $\vec{x}_{r}$ to $\vec{x}_{r}^{i}$ and $\vec{x}_{r}^{o}$ \textup{, Equation \ref{eq:radial_clust}}\\
\textup{Find} $\vec{\xi}$ \textup{and} $\alpha(\vec{\xi})$ \textup{, Equations \ref{eq:cost} and \ref{eq:huber}} \\
\textup{Step} $\Gamma$
}
\textup{Find} $r_{m}$ and $R_{m}$, Equation \ref{eq:rm} \\
\SetKw{Inspect}{Inspect}
\Inspect $Z'$, Equation \ref{eq:z_csk}
}
\end{algorithm}

\begin{figure}[ht]
\center
\includegraphics[scale=0.325]{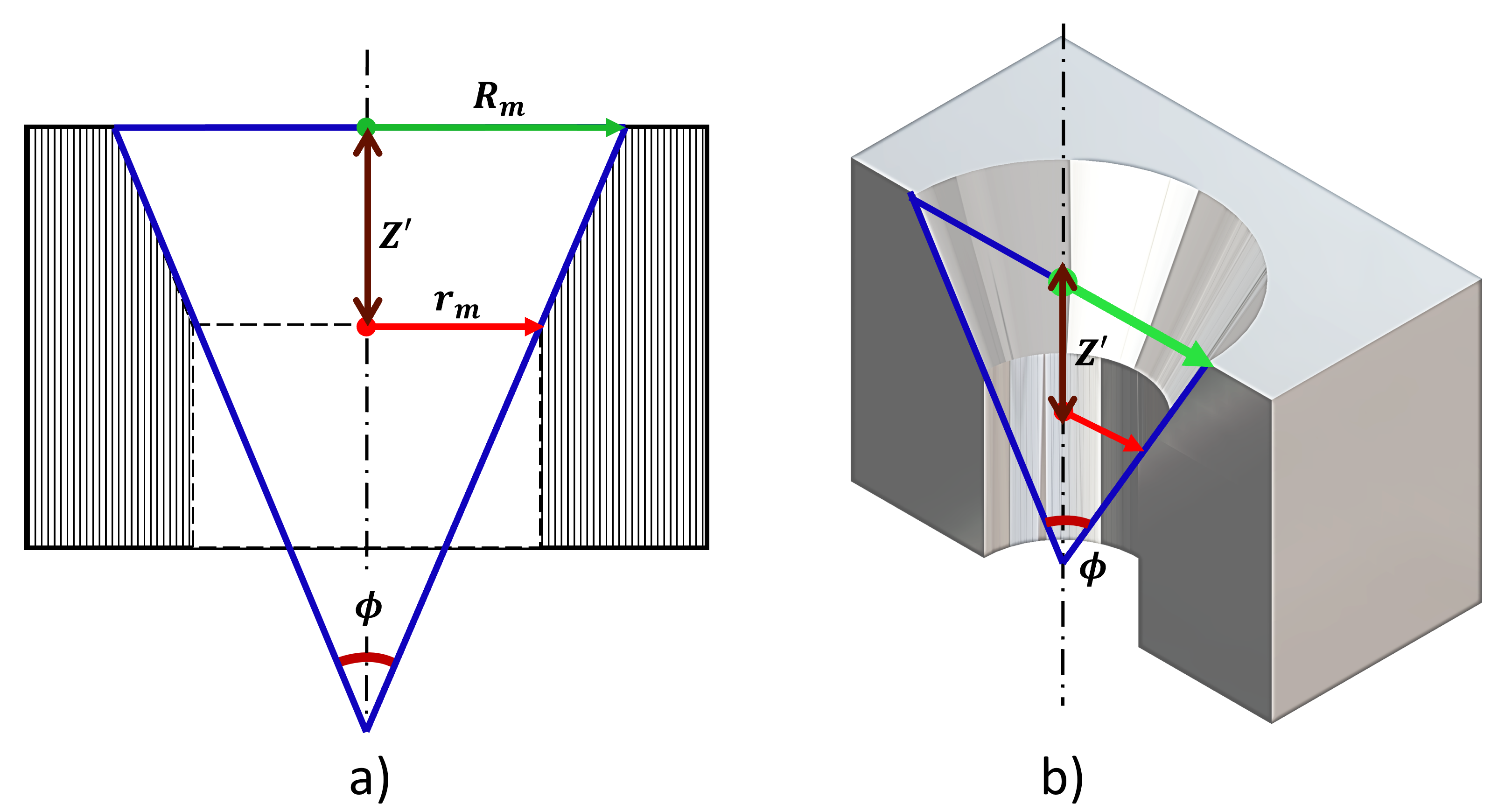}
\caption{a) Front and b) isometric views of the. The  depth $Z'$ can be derived by trigonometry, given $r_{m}$, $R_{m}$, and $\phi$.}
\label{fig:csk_fig}
\end{figure}

Having the inner and outer radii, a known angle $\phi$, and $Z$, the  depth $Z'$ is estimated by simple trigonometry. Figure \ref{fig:csk_fig} presents front and isometric views of the s for a visual illustration of the trigonometric relation. First, the inner and outer radii are converted from pixels to mm by

\begin{equation}
    \begin{bmatrix}
        r_{m} \\
        R_{m}
    \end{bmatrix} = \begin{bmatrix}
        \frac{Z}{F} & 0 \\
        0 & \frac{Z}{F}
    \end{bmatrix} \begin{bmatrix}
        r \\
        R
    \end{bmatrix}
    \label{eq:rm}
\end{equation}

\noindent where $r_{m}$ and $R_{m}$ are the radii expressed in mm. Intuitively, $Z'$ can be inferred from Figure \ref{fig:csk_fig} and is expressed by

\begin{equation}
    Z' = \frac{R_{m}- r_{m}}{\tan\frac{\phi}{2}}.
    \label{eq:z_csk}
\end{equation}\

\begin{figure*}[ht!]
\center
\includegraphics[width=0.9\textwidth,height=0.5\textheight,scale=1]{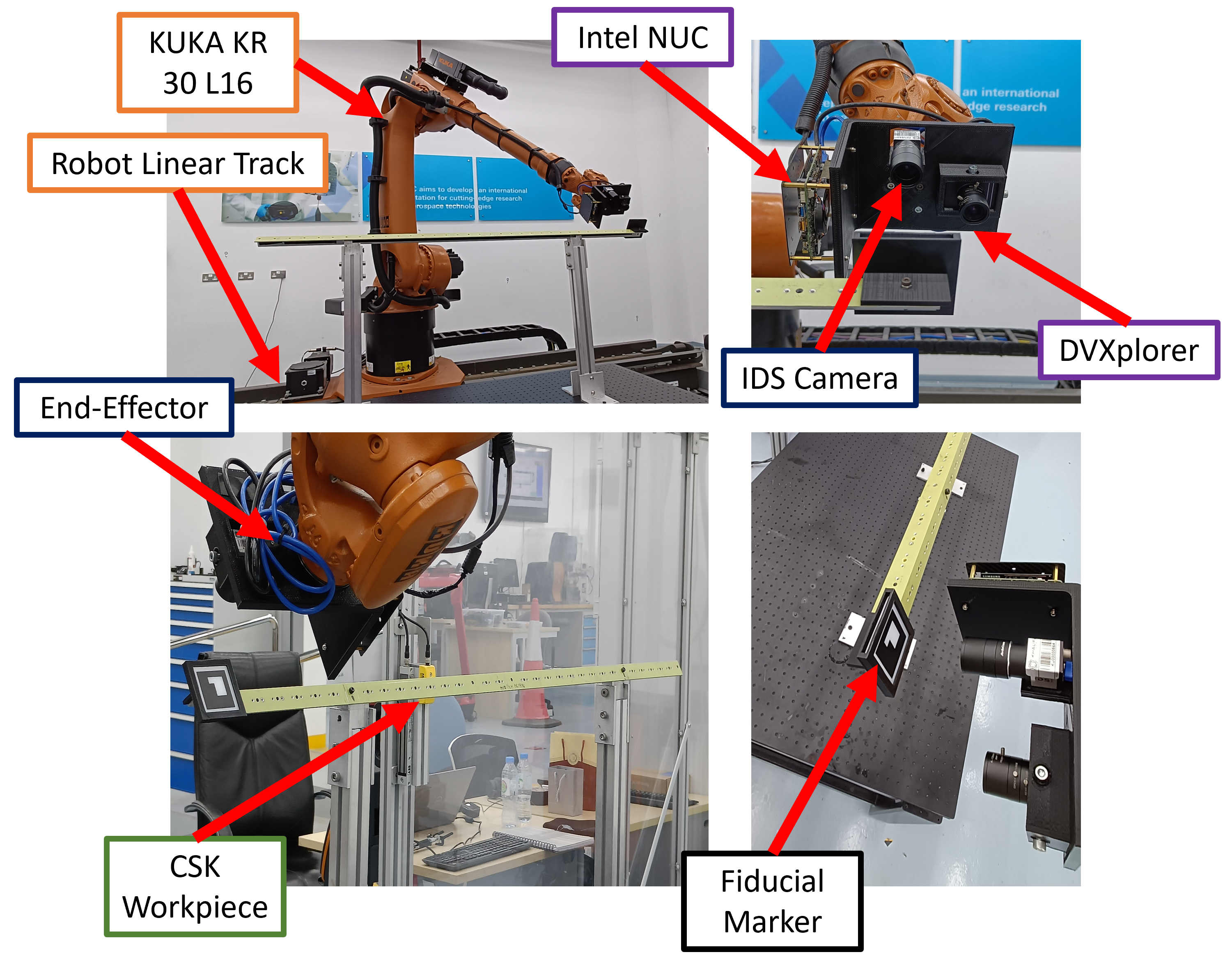}
\caption{Experimental setup for validating the proposed  inspection framework.  A KUKA robotic manipulator is utilized for moving the neuromorphic camera to inspect a countersinks workpiece representing large industrial parts. A fiducial marker is installed on the workpiece for the normal alignment of the sensor with the workpiece.}
\label{fig:exp_setup}
\end{figure*}

\section{Experimental Validation and Results} \label{sec:exps}
\subsection{Experimental Setup}
\begin{figure}[ht]
\center
\includegraphics[scale=0.275]{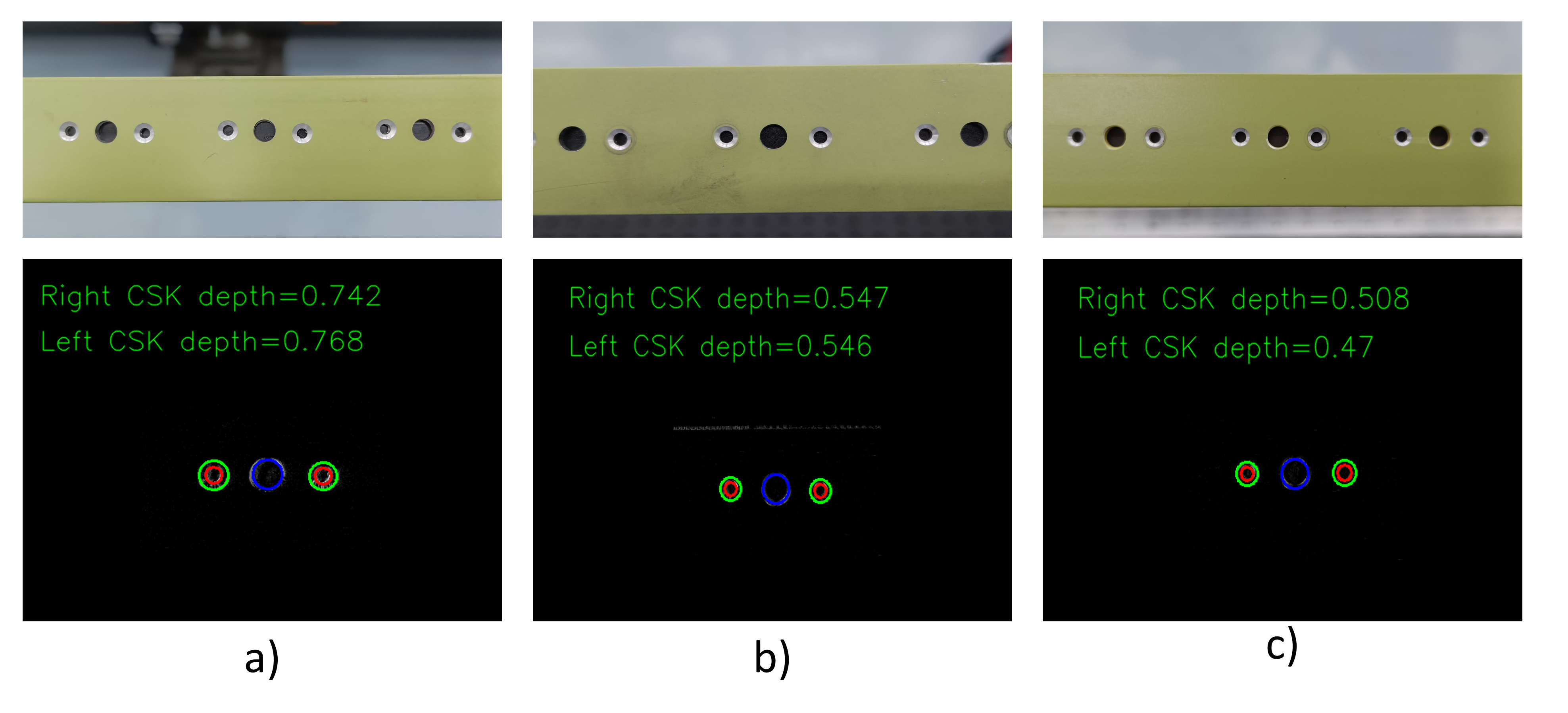}
\caption{Inspected workpieces with countersinks of varying depth specifications along with the inspected specimens by our method. a)-c) are the workpiece IDs ordered in descending order with respect to the nominal countersink depths.}
\label{fig:csk_workpiece}
\end{figure}

We have tested our framework in a variety of challenging scenarios. Fig. \ref{fig:exp_setup} shows the designed experimental setup to validate the proposed approach. A 7-DOF KUKA KR30 L16 with KL-1000-2 linear track was utilized to move the NVS during the inspection, and a workpiece resembling large indsutrial structures with inspection requirement. The robot end-effector comprises of IDS UI-5860CP ($1920 \times 1080$) as the frame-based sensor for the normal alignment clarified in section \ref{sec:overview}, while the DVXplorer ($640 \times 480$) as the NVS. Finally, Intel® NUC5i5RYH was also mounted on the end-effector for onboard processing, where both imaging sensors were interfaced with the robot operating system (ROS) using a USB 3.0 terminal. Four important aspects of the experimental protocol need to be taken into consideration. First, the NVS motion during the inspection ranged from 0.05 m/s to 0.5 m/s to justify the purpose of this paper. Second, the NVS moved in 2D linear motion as it sweeps the countersinks workpiece, as discussed in section \ref{sec:detection}. Third, we performed the proposed inspection method on three workpiece variants of different countersink sizes to show that the work of this paper is generalizable to different dimensions of countersink specimens, see Fig. \ref{fig:csk_workpiece}. Finally, the standoff between the NVS and the workpiece during the sweep was $90 mm$ to maintain high inspection precision. Videos of the experiments are available through the following link \url{https://www.dropbox.com/s/pateqqwh4d605t3/final_video_new.mp4?dl=0}.

The experimental validation of the proposed work revolves around three folds. In section \ref{sec:performance}, we compare the warping-Huber-based circle detection performance in terms of detection success rate and the IWEs image quality. We also compare the aforementioned detector to ED-Circles on frames from IDS camera and E2VID \cite{e2vid} network for event-based image reconstruction. In section \ref{sec:depth_precision}, we evaluate the precision and repeatability of the  inspection by computing the standard deviation of the estimated  depths on 10 runs at each NVS speed. Finally, section \ref{sec:benchmarks} benchmarks the work of this paper to state-of-the-art methods for inspection in terms of precision and inspection execution time.

\subsection{Countersink Detection Performance} \label{sec:performance}
\begin{figure}[t]
\center
\includegraphics[scale=0.335]{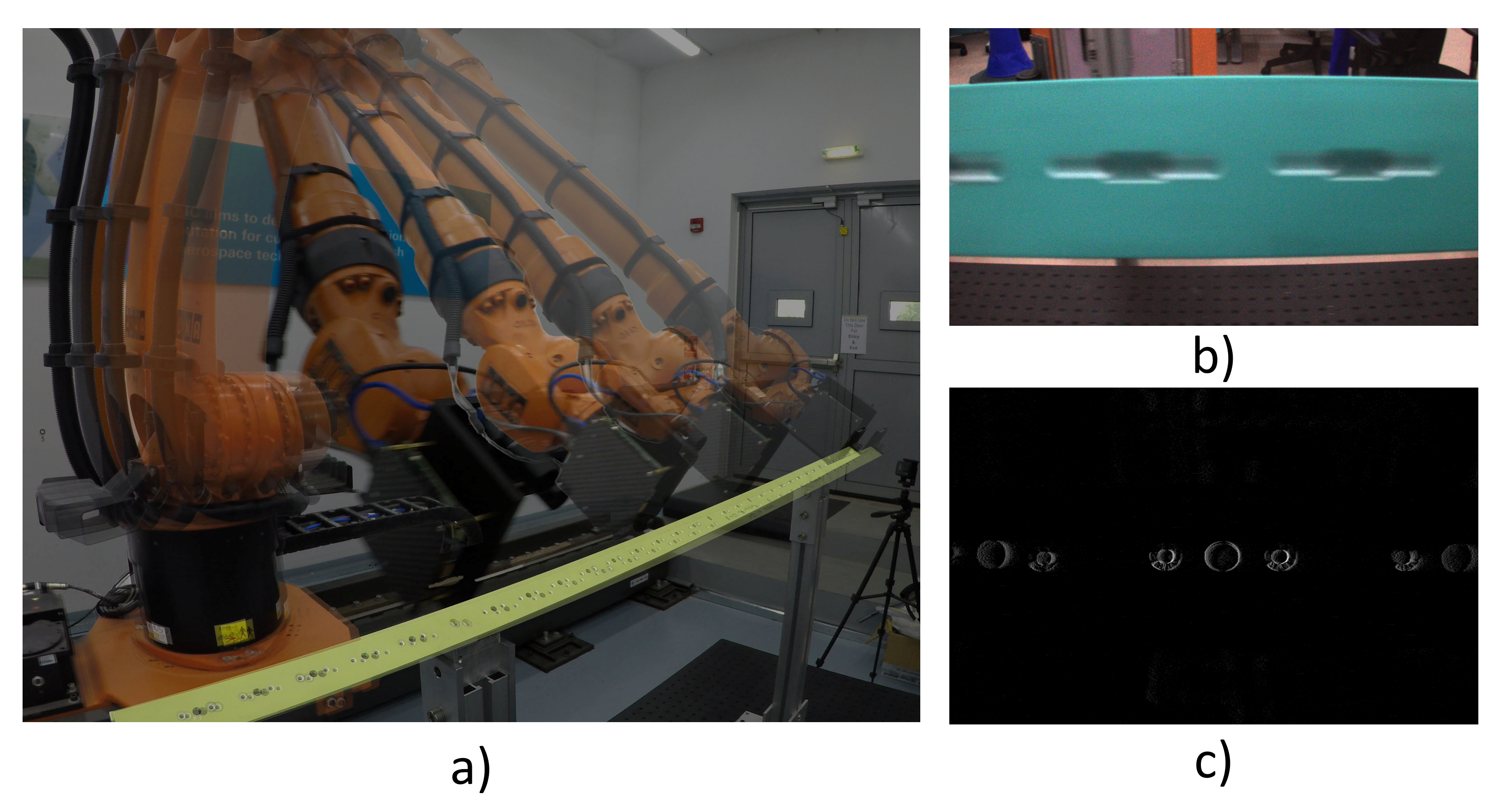}
\caption{a) NVS motion trajectory during the inspection, b) and c) sharp IWEs and IDS frames suffering from motion blur. The reconstructed IWEs justify the purpose of the NVS for high-speed inspection.}
\label{fig:kuka_burst}
\end{figure}

\begin{figure*}[ht!]
\centering
\begin{tabular}{m{0.065\textwidth}|>{\centering\arraybackslash}m{0.2\textwidth}|>{\centering\arraybackslash}m{0.2\textwidth}|>{\centering\arraybackslash}m{0.2\textwidth}|>{\centering\arraybackslash}m{0.2\textwidth}}
 & (a) Accumulated Event Frames \(\Delta t = 33 ms\)  & (b) ED-Circles on IDS Frames & (c) ED-Circles on E2VID Reconstructed Images & (d) Proposed Detector \\ \hline
 0.05 m/s NVS Speed & 
 \includegraphics[width=0.2\textwidth, height=0.12\textwidth]{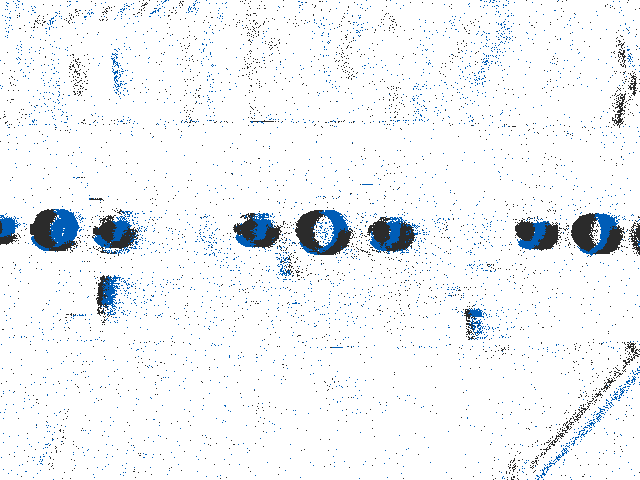} & 
 \includegraphics[width=0.2\textwidth]{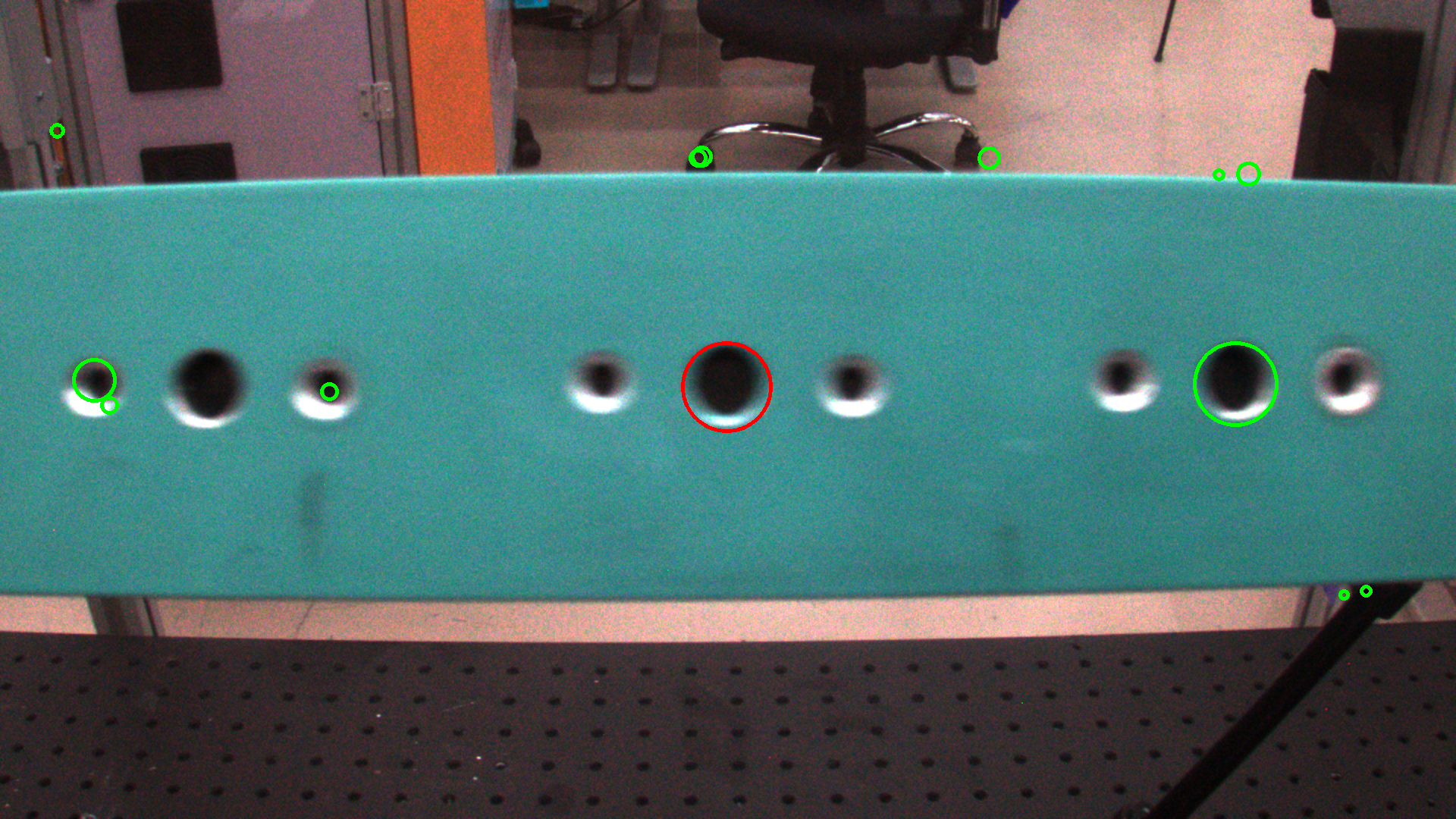} &
 \includegraphics[width=0.2\textwidth]{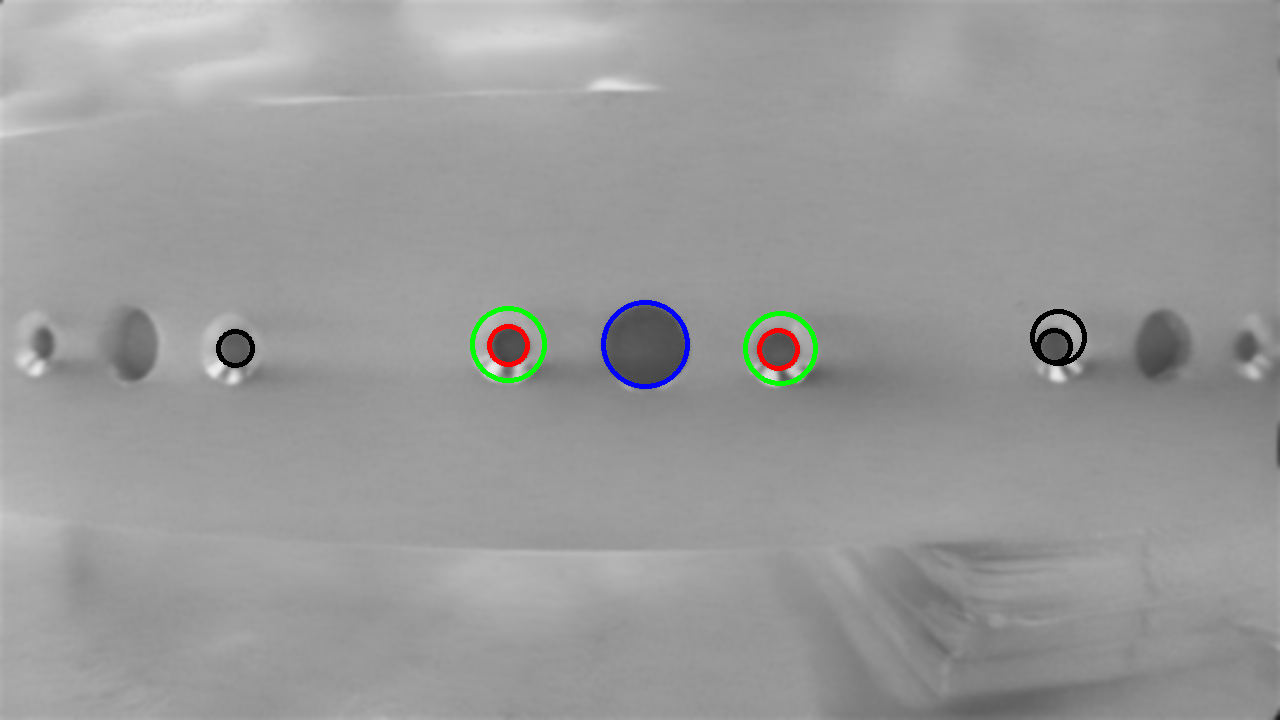} &
 \includegraphics[width=0.2\textwidth]{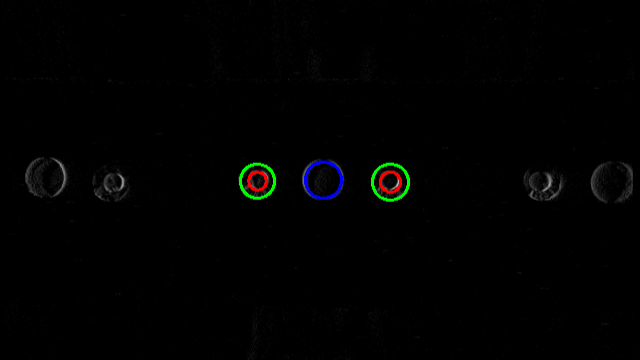} 
 \\ \hline
 0.5 m/s NVS Speed &
 \includegraphics[width=0.2\textwidth, height=0.12\textwidth]{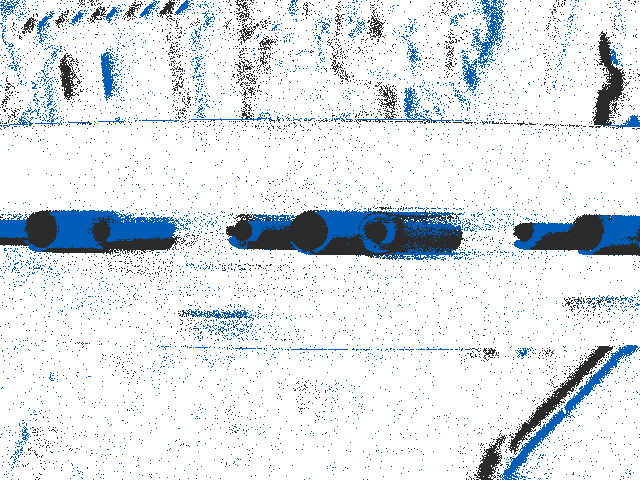} & 
 \includegraphics[width=0.2\textwidth]{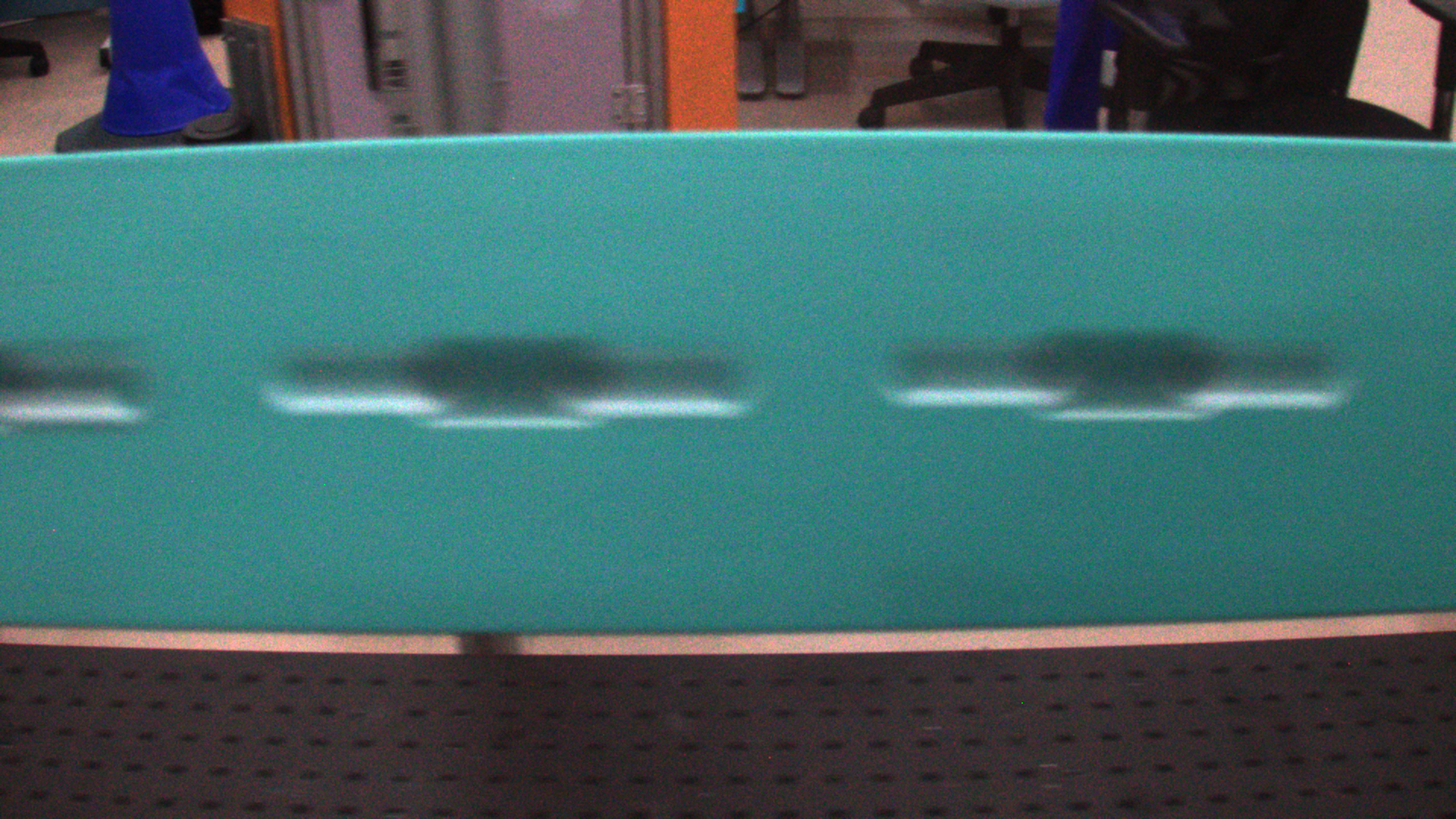} &
 \includegraphics[width=0.2\textwidth]{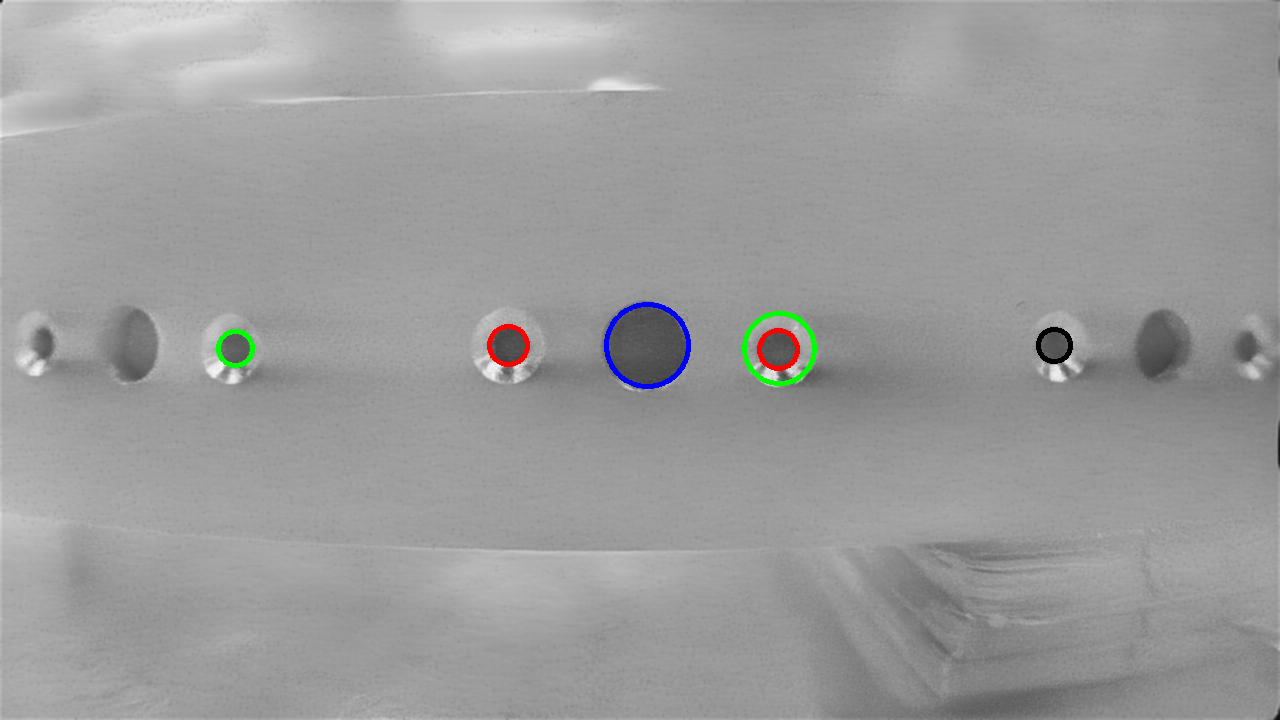} &
 \includegraphics[width=0.2\textwidth]{figures/robust_huber_2.png}
 \\
 \multicolumn{5}{c}{\includegraphics[width=0.25\textwidth]{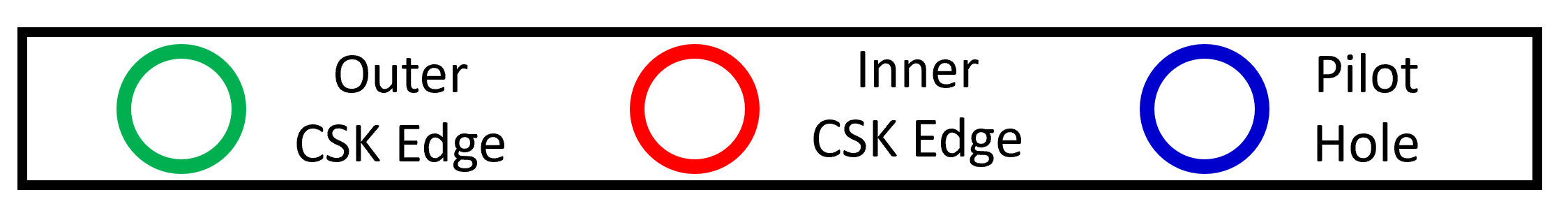}} 
 \\ 
\end{tabular}
\caption{Qualitative comparison of the d) proposed detector with a) artificial event frames accumulated at $\Delta t=33ms$, b) conventional frames from the IDS camera, and c) E2VID reconstructed grayscale images. Notice that the  features in the artificial event frames are perturbed by motion blur and noise, while features IDS images are degraded by motion blur even. On the other hand, E2VID provides textured image reconstruction of the scene, but ED-Circles fails due to reflections. Our detector, on the other hand, shows improved performance at high speeds with robustness against noise, reflection outliers, and motion blur.}
\label{fig:circle_fixed_iwe}
\end{figure*}

\begin{figure*}[ht]
\center
\includegraphics[width=\textwidth,height=0.125\textheight,scale=1]{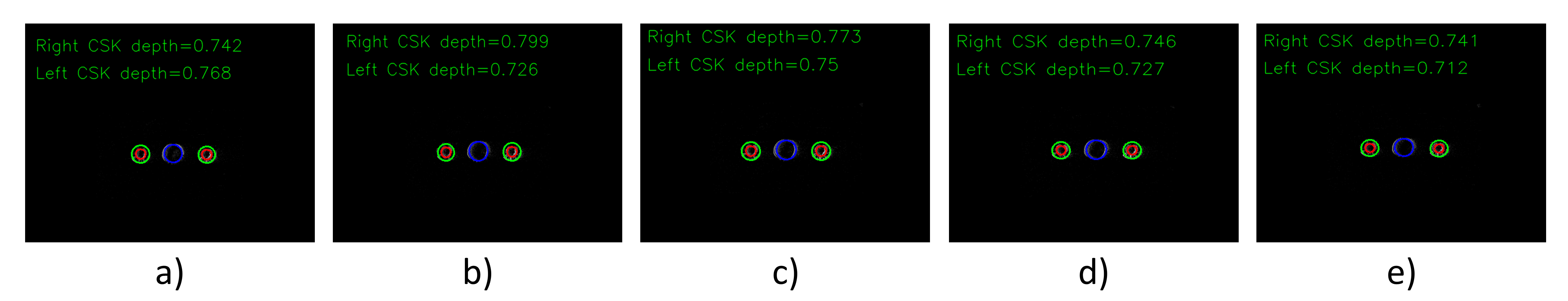}
\caption{a)-e) Shows the precision of the proposed robust circle detection and depth estimation algorithms at 0.5 m/s. Notice the precise measurements even though the NVS resolution is relatively low.}
\label{fig:csk_txt}
\end{figure*}

The quality of the reconstructed IWEs directly impacts the  detection performance. Figure \ref{fig:kuka_burst} compares the obtained neuromorphic IWEs by warping against conventional images for a sweep speed of 0.5 m/s. If artificial event frames are formulated by simple accumulation at fixed time intervals, motion blur is witnessed, and the detection performance is highly influenced by sensor noise, see Figure \ref{fig:circle_fixed_iwe}. Given the improved quality of IWEs, the mean shift clustering algorithm for  detection is evaluated based on its detection success rate, as a detection rate of 100\% was obtained for speeds up to 0.5 m/s.

We also provide a qualitative comparison of the IWEs and the proposed circle detector in section \ref{sec:depth} against ED-Circles applied on IDS conventional frames and reconstructed images from the E2VID \cite{e2vid} network, illustrated in Figure \ref{fig:circle_fixed_iwe}. At 0.05 m/s, ED-Circles on IDS images fail to detect the circular edges of the countersinks due to motion blur, while detection was successful on E2VID network. This is because the network is characterized by its capability of reconstructing deblurred images of the scene. However, ED-Circles fail to detect the holes due to the inner wall reflections reducing the gradient of the  circular edges at 0.5 m/s. In addition, E2VID requires a powerful GPU to run online, which is difficult to install on onboard processors. Contrary to the aforementioned approaches, the proposed detection pipeline runs on a low-end Intel® Core™ i5-5250U CPU, leverages the asynchronous nature of neuromorphic events, and demonstrates robustness against motion blur and outliers.

\begin{table*}[ht]
\centering
\caption{Computation of standard deviation in mm for sample holes at a designated NVS speed for 10 trials. a)-c) represent the workpiece ID.}
\resizebox{\textwidth}{!}{%
\begin{tabular}{|c|cccccccccc|c|}
\hline
\multirow{2}{*}{\textbf{Hole ID}} & \multicolumn{10}{c|}{\textbf{Trial Number}} & \multirow{2}{*}{$\boldsymbol{\sigma_{d}}$} \\ \cline{2-11}
 & \multicolumn{1}{c|}{\textbf{1}} & \multicolumn{1}{c|}{\textbf{2}} & \multicolumn{1}{c|}{\textbf{3}} & \multicolumn{1}{c|}{\textbf{4}} & \multicolumn{1}{c|}{\textbf{5}} & \multicolumn{1}{c|}{\textbf{6}} & \multicolumn{1}{c|}{\textbf{7}} & \multicolumn{1}{c|}{\textbf{8}} & \multicolumn{1}{c|}{\textbf{9}} & \textbf{10} &  \\ \hline
\textbf{1a} & \multicolumn{1}{c|}{0.728} & \multicolumn{1}{c|}{0.829} & \multicolumn{1}{c|}{0.777} & \multicolumn{1}{c|}{0.800} & \multicolumn{1}{c|}{0.790} & \multicolumn{1}{c|}{0.813} & \multicolumn{1}{c|}{0.771} & \multicolumn{1}{c|}{0.823} & \multicolumn{1}{c|}{0.830} & 0.773 & 0.031 \\ \hline
\textbf{1b} & \multicolumn{1}{c|}{0.557} & \multicolumn{1}{c|}{0.574} & \multicolumn{1}{c|}{0.570} & \multicolumn{1}{c|}{0.544} & \multicolumn{1}{c|}{0.511} & \multicolumn{1}{c|}{0.526} & \multicolumn{1}{c|}{0.545} & \multicolumn{1}{c|}{0.532} & \multicolumn{1}{c|}{0.584} & 0.534 & 0.025 \\ \hline
\textbf{1c} & \multicolumn{1}{c|}{0.481} & \multicolumn{1}{c|}{0.470} & \multicolumn{1}{c|}{0.439} & \multicolumn{1}{c|}{0.463} & \multicolumn{1}{c|}{0.483} & \multicolumn{1}{c|}{0.439} & \multicolumn{1}{c|}{0.441} & \multicolumn{1}{c|}{0.462} & \multicolumn{1}{c|}{0.467} & 0.473 & 0.016 \\ \hline
\end{tabular}%
}
\label{table:precision_samples}
\end{table*}

\begin{table*}[ht]
\centering
\caption{Precision of the countersink inspection system in mm for 10 trials at each NVS speed for workpiece with ID 'A', where a precision of 0.0275 mm is achieved.}
\resizebox{\textwidth}{!}{%
\begin{tabular}{ccccccccccc|}
\hline
\multicolumn{1}{|c|}{\multirow{2}{*}{\textbf{\begin{tabular}[c]{@{}c@{}}Speed\\ (m/s)\end{tabular}}}} & \multicolumn{10}{c|}{\textbf{Hole ID}} \\ \cline{2-11} 
\multicolumn{1}{|c|}{} & \multicolumn{1}{c|}{\textbf{1}} & \multicolumn{1}{c|}{\textbf{2}} & \multicolumn{1}{c|}{\textbf{3}} & \multicolumn{1}{c|}{\textbf{4}} & \multicolumn{1}{c|}{\textbf{5}} & \multicolumn{1}{c|}{\textbf{6}} & \multicolumn{1}{c|}{\textbf{7}} & \multicolumn{1}{c|}{\textbf{8}} & \multicolumn{1}{c|}{\textbf{9}} & \textbf{10} \\ \hline
\multicolumn{1}{|c|}{\textbf{0.05}} & \multicolumn{1}{c|}{0.021} & \multicolumn{1}{c|}{0.020} & \multicolumn{1}{c|}{0.028} & \multicolumn{1}{c|}{0.022} & \multicolumn{1}{c|}{0.031} & \multicolumn{1}{c|}{0.029} & \multicolumn{1}{c|}{0.035} & \multicolumn{1}{c|}{0.017} & \multicolumn{1}{c|}{0.011} & 0.029 \\ \hline
\multicolumn{1}{|c|}{\textbf{0.1}} & \multicolumn{1}{c|}{0.016} & \multicolumn{1}{c|}{0.023} & \multicolumn{1}{c|}{0.011} & \multicolumn{1}{c|}{0.023} & \multicolumn{1}{c|}{0.034} & \multicolumn{1}{c|}{0.035} & \multicolumn{1}{c|}{0.033} & \multicolumn{1}{c|}{0.036} & \multicolumn{1}{c|}{0.032} & 0.035 \\ \hline
\multicolumn{1}{|c|}{\textbf{0.2}} & \multicolumn{1}{c|}{0.022} & \multicolumn{1}{c|}{0.018} & \multicolumn{1}{c|}{0.016} & \multicolumn{1}{c|}{0.025} & \multicolumn{1}{c|}{0.028} & \multicolumn{1}{c|}{0.031} & \multicolumn{1}{c|}{0.026} & \multicolumn{1}{c|}{0.025} & \multicolumn{1}{c|}{0.029} & 0.028 \\ \hline
\multicolumn{1}{|c|}{\textbf{0.3}} & \multicolumn{1}{c|}{0.023} & \multicolumn{1}{c|}{0.016} & \multicolumn{1}{c|}{0.021} & \multicolumn{1}{c|}{0.029} & \multicolumn{1}{c|}{0.033} & \multicolumn{1}{c|}{0.026} & \multicolumn{1}{c|}{0.021} & \multicolumn{1}{c|}{0.033} & \multicolumn{1}{c|}{0.032} & 0.029 \\ \hline
\multicolumn{1}{|c|}{\textbf{0.5}} & \multicolumn{1}{c|}{0.031} & \multicolumn{1}{c|}{0.031} & \multicolumn{1}{c|}{0.030} & \multicolumn{1}{c|}{0.033} & \multicolumn{1}{c|}{0.027} & \multicolumn{1}{c|}{0.032} & \multicolumn{1}{c|}{0.025} & \multicolumn{1}{c|}{0.019} & \multicolumn{1}{c|}{0.033} & 0.026 \\ \hline
\multicolumn{1}{|c|}{\boldsymbol{$\sigma_{r}$}} & \multicolumn{1}{c|}{0.0209} & \multicolumn{1}{c|}{0.022} & \multicolumn{1}{c|}{0.0224} & \multicolumn{1}{c|}{0.0267} & \multicolumn{1}{c|}{0.0334} & \multicolumn{1}{c|}{0.0307} & \multicolumn{1}{c|}{0.0315} & \multicolumn{1}{c|}{0.0271} & \multicolumn{1}{c|}{0.0286} & 0.0296 \\ \hline
 &  &  &  &  &  &  &  & \multicolumn{1}{c|}{} & \multicolumn{1}{c|}{\textbf{Aggregate}} & \textbf{0.0275} \\ \cline{10-11} 
\end{tabular}%
}
\label{table:precision1}
\end{table*}

\begin{table*}[ht]
\centering
\caption{Precision of the countersink inspection system in mm for 10 trials at each NVS speed for workpiece with ID 'B', where a precision of 0.0251 mm is achieved.}
\resizebox{\textwidth}{!}{%
\begin{tabular}{ccccccccccc|}
\hline
\multicolumn{1}{|c|}{\multirow{2}{*}{\textbf{\begin{tabular}[c]{@{}c@{}}Speed\\ (m/s)\end{tabular}}}} & \multicolumn{10}{c|}{\textbf{Hole ID}} \\ \cline{2-11} 
\multicolumn{1}{|c|}{} & \multicolumn{1}{c|}{\textbf{1}} & \multicolumn{1}{c|}{\textbf{2}} & \multicolumn{1}{c|}{\textbf{3}} & \multicolumn{1}{c|}{\textbf{4}} & \multicolumn{1}{c|}{\textbf{5}} & \multicolumn{1}{c|}{\textbf{6}} & \multicolumn{1}{c|}{\textbf{7}} & \multicolumn{1}{c|}{\textbf{8}} & \multicolumn{1}{c|}{\textbf{9}} & \textbf{10} \\ \hline
\multicolumn{1}{|c|}{\textit{\textbf{0.05}}} & \multicolumn{1}{c|}{0.022} & \multicolumn{1}{c|}{0.022} & \multicolumn{1}{c|}{0.019} & \multicolumn{1}{c|}{0.021} & \multicolumn{1}{c|}{0.033} & \multicolumn{1}{c|}{0.014} & \multicolumn{1}{c|}{0.043} & \multicolumn{1}{c|}{0.019} & \multicolumn{1}{c|}{0.014} & 0.024 \\ \hline
\multicolumn{1}{|c|}{\textit{\textbf{0.1}}} & \multicolumn{1}{c|}{0.019} & \multicolumn{1}{c|}{0.042} & \multicolumn{1}{c|}{0.017} & \multicolumn{1}{c|}{0.020} & \multicolumn{1}{c|}{0.023} & \multicolumn{1}{c|}{0.025} & \multicolumn{1}{c|}{0.023} & \multicolumn{1}{c|}{0.023} & \multicolumn{1}{c|}{0.024} & 0.018 \\ \hline
\multicolumn{1}{|c|}{\textit{\textbf{0.2}}} & \multicolumn{1}{c|}{0.018} & \multicolumn{1}{c|}{0.027} & \multicolumn{1}{c|}{0.019} & \multicolumn{1}{c|}{0.011} & \multicolumn{1}{c|}{0.022} & \multicolumn{1}{c|}{0.027} & \multicolumn{1}{c|}{0.036} & \multicolumn{1}{c|}{0.022} & \multicolumn{1}{c|}{0.027} & 0.027 \\ \hline
\multicolumn{1}{|c|}{\textit{\textbf{0.3}}} & \multicolumn{1}{c|}{0.027} & \multicolumn{1}{c|}{0.017} & \multicolumn{1}{c|}{0.032} & \multicolumn{1}{c|}{0.022} & \multicolumn{1}{c|}{0.027} & \multicolumn{1}{c|}{0.032} & \multicolumn{1}{c|}{0.033} & \multicolumn{1}{c|}{0.024} & \multicolumn{1}{c|}{0.026} & 0.024 \\ \hline
\multicolumn{1}{|c|}{\textit{\textbf{0.5}}} & \multicolumn{1}{c|}{0.025} & \multicolumn{1}{c|}{0.019} & \multicolumn{1}{c|}{0.030} & \multicolumn{1}{c|}{0.023} & \multicolumn{1}{c|}{0.021} & \multicolumn{1}{c|}{0.011} & \multicolumn{1}{c|}{0.029} & \multicolumn{1}{c|}{0.016} & \multicolumn{1}{c|}{0.031} & 0.023 \\ \hline
\multicolumn{1}{|c|}{$\boldsymbol{\sigma_{r}}$} & \multicolumn{1}{c|}{0.0261} & \multicolumn{1}{c|}{0.0269} & \multicolumn{1}{c|}{0.0242} & \multicolumn{1}{c|}{0.0190} & \multicolumn{1}{c|}{0.0256} & \multicolumn{1}{c|}{0.0232} & \multicolumn{1}{c|}{0.0335} & \multicolumn{1}{c|}{0.0210} & \multicolumn{1}{c|}{0.0251} & 0.0234 \\ \hline
 &  &  &  &  &  &  &  & \multicolumn{1}{c|}{} & \multicolumn{1}{c|}{\textbf{Aggregate}} & 0.0251 \\ \cline{10-11} 
\end{tabular}%
}
\label{table:precision2}
\end{table*}

\begin{table*}[ht]
\centering
\caption{Precision of the countersink inspection system in mm for 10 trials at each NVS speed for workpiece with ID 'C', where a precision of 0.0240 mm is achieved.}
\resizebox{\textwidth}{!}{%
\begin{tabular}{ccccccccccc|}
\hline
\multicolumn{1}{|c|}{\multirow{2}{*}{\textbf{\begin{tabular}[c]{@{}c@{}}Speed\\ (m/s)\end{tabular}}}} & \multicolumn{10}{c|}{\textbf{Hole ID}} \\ \cline{2-11} 
\multicolumn{1}{|c|}{} & \multicolumn{1}{c|}{\textbf{1}} & \multicolumn{1}{c|}{\textbf{2}} & \multicolumn{1}{c|}{\textbf{3}} & \multicolumn{1}{c|}{\textbf{4}} & \multicolumn{1}{c|}{\textbf{5}} & \multicolumn{1}{c|}{\textbf{6}} & \multicolumn{1}{c|}{\textbf{7}} & \multicolumn{1}{c|}{\textbf{8}} & \multicolumn{1}{c|}{\textbf{9}} & \textbf{10} \\ \hline
\multicolumn{1}{|c|}{\textit{\textbf{0.05}}} & \multicolumn{1}{c|}{0.017} & \multicolumn{1}{c|}{0.024} & \multicolumn{1}{c|}{0.033} & \multicolumn{1}{c|}{0.024} & \multicolumn{1}{c|}{0.024} & \multicolumn{1}{c|}{0.027} & \multicolumn{1}{c|}{0.021} & \multicolumn{1}{c|}{0.020} & \multicolumn{1}{c|}{0.034} & 0.019 \\ \hline
\multicolumn{1}{|c|}{\textit{\textbf{0.1}}} & \multicolumn{1}{c|}{0.026} & \multicolumn{1}{c|}{0.018} & \multicolumn{1}{c|}{0.021} & \multicolumn{1}{c|}{0.019} & \multicolumn{1}{c|}{0.012} & \multicolumn{1}{c|}{0.029} & \multicolumn{1}{c|}{0.018} & \multicolumn{1}{c|}{0.019} & \multicolumn{1}{c|}{0.016} & 0.021 \\ \hline
\multicolumn{1}{|c|}{\textit{\textbf{0.2}}} & \multicolumn{1}{c|}{0.023} & \multicolumn{1}{c|}{0.022} & \multicolumn{1}{c|}{0.016} & \multicolumn{1}{c|}{0.013} & \multicolumn{1}{c|}{0.017} & \multicolumn{1}{c|}{0.030} & \multicolumn{1}{c|}{0.016} & \multicolumn{1}{c|}{0.027} & \multicolumn{1}{c|}{0.026} & 0.027 \\ \hline
\multicolumn{1}{|c|}{\textit{\textbf{0.3}}} & \multicolumn{1}{c|}{0.013} & \multicolumn{1}{c|}{0.011} & \multicolumn{1}{c|}{0.027} & \multicolumn{1}{c|}{0.011} & \multicolumn{1}{c|}{0.019} & \multicolumn{1}{c|}{0.023} & \multicolumn{1}{c|}{0.034} & \multicolumn{1}{c|}{0.029} & \multicolumn{1}{c|}{0.027} & 0.027 \\ \hline
\multicolumn{1}{|c|}{\textit{\textbf{0.5}}} & \multicolumn{1}{c|}{0.016} & \multicolumn{1}{c|}{0.012} & \multicolumn{1}{c|}{0.019} & \multicolumn{1}{c|}{0.021} & \multicolumn{1}{c|}{0.020} & \multicolumn{1}{c|}{0.021} & \multicolumn{1}{c|}{0.032} & \multicolumn{1}{c|}{0.037} & \multicolumn{1}{c|}{0.028} & 0.029 \\ \hline
\multicolumn{1}{|c|}{$\boldsymbol{\sigma_{r}}$} & \multicolumn{1}{c|}{0.0219} & \multicolumn{1}{c|}{0.0182} & \multicolumn{1}{c|}{0.0268} & \multicolumn{1}{c|}{0.0183} & \multicolumn{1}{c|}{0.0188} & \multicolumn{1}{c|}{0.0262} & \multicolumn{1}{c|}{0.0253} & \multicolumn{1}{c|}{0.0272} & \multicolumn{1}{c|}{0.0293} & 0.0249 \\ \hline
 &  &  &  &  &  &  &  & \multicolumn{1}{c|}{} & \multicolumn{1}{c|}{\textbf{Aggregate}} & 0.0240 \\ \cline{10-11} 
\end{tabular}%
}
\label{table:precision3}
\end{table*}

\subsection{Precision of Countersink Depth Estimation} \label{sec:depth_precision}
While the proposed approach provides the required detection success rates, the precision of the  depth estimation needs to be evaluated. The standard metric for evaluating the inspection is the standard deviation $\sigma_{d}$, where $\sigma_{d}$ was evaluated on each hole of each workpiece for 10 runs at distinct speeds of NVS. Table \ref{table:precision_samples} shows the calculation $\sigma_{d}$ for a countersink sample from each workpiece at a speed of 0.5 m/s for 10 trials, and Figure \ref{fig:csk_txt} visualizes these estimations. The precision of the proposed inspection pipeline is directly correlated with the performance of the robust circle detection algorithm discussed in section \ref{sec:depth}. Notice that the radii of the fitted circles are consistent, where $\sigma_{d}$ = 0.031 mm was obtained. Tables \ref{table:precision1}, \ref{table:precision2}, \ref{table:precision3} reports the precision of all $\sigma_{d}$ on each workpiece, where the proposed method provides an aggregate inspection precision of 0.025 mm.

Notice that the countersink inspection precision is higher for specimens with lower depths. This is because the motion compensation performance tends to be improved as the depth of the inner countersink walls is closer to the utilized depth for warping the events, as discussed in section \ref{sec:detection}. In all the cases however, the proposed inspection framework achieves a precision of 0.025 mm, abiding by the industrial requirements of various assembly lines. The reason behind the high precision of the proposed circle detector, even on a low-resolution sensor, is that it relies on an optimization approach fitting the best candidate circle to the observed hole edges. The high dependence of conventional circle detectors on the image gradient degrades their performance even on high-resolution cameras. Assuming that the conventional imaging sensor is static during the inspection and no motion blur is witnessed, the lighting conditions of the inspection setup have a major influence in controlling the precision of these circle detectors. In contrast, the high dynamic range of the NVS makes our method robust against illumination variations.

\begin{table*}[ht!]
\centering
\caption{Quantitative comparison of the proposed framework against state-of-the-art works for  inspection using monocular vision. Even with a much lower resolution sensor, our method provides comparable results with the same order of magnitude against state-of-the-art approaches. Note that \Cross \text{} indicates that the metric is unavailable.}
\resizebox{\textwidth}{!}{%
\begin{tabular}{|c|c|c|c|c|c|}
\hline
\textbf{Authors} & \textbf{Sensor Technology} & \textbf{Year} & \textbf{Sensor Resolution} & \textbf{Precision (mm)} & \textbf{Inspection Time (s)} \\ \hline
R. Haldiman \cite{3d_csk_measurement} & DLP Projector & 2015 & \Cross & 0.03 & \Cross \\ \hline
Yu et al. \cite{yu2019vision} & Conventional Camera & 2019 & 2050 x 2448 & 0.02 & 42 \\ \hline
\textbf{Ours} & \textbf{NVS} & \textbf{2023} & \textbf{640 x 480} & \textbf{0.025} & \textbf{4.98} \\ \hline
\end{tabular}%
}
\label{table:benchmarks}
\end{table*}

\subsection{Benchmarks} \label{sec:benchmarks}
To evaluate the impact of our event-driven approach, we compare the obtained results against state-of-the-art vision-based methods for  inspection, which comprise of Haldiman's \cite{3d_csk_measurement} inspection using Digital Light Processing (DLP) projector and Yu et al. work \cite{sa2018design}. These works are compared quantitatively in terms of precision and qualitatively in terms of inspection speed, where the comparison is more elaborated in Table \ref{table:benchmarks}.

Our approach shows comparable results with the same orders of magnitude compared to previous works. The work of this paper outperforms Haldiman' \cite{3d_csk_measurement} with the advantage of utilizing a neuromorphic sensor without the need for an active light projector. More importantly, the usage of laser projectors for depth estimation is currently unfavorable in industries for safety precautions, especially with the introduction of human-robot interaction in assembly lines. Yu et al. \cite{yu2019vision}, on the other hand, outperforms our method by 0.0125 mm. Nevertheless, our method provides high-speed inspection and meets current industrial standards. Furthermore, the resolution of their utilized sensor is much higher than the DVXplorer by a factor of more than $1/16$. With the current advancements in neuromorphic sensor models, high-definition neuromorphic cameras easily circumvent this limitation and can provide improved precision.

\begin{figure}[ht]
\center
\includegraphics[scale=0.25]{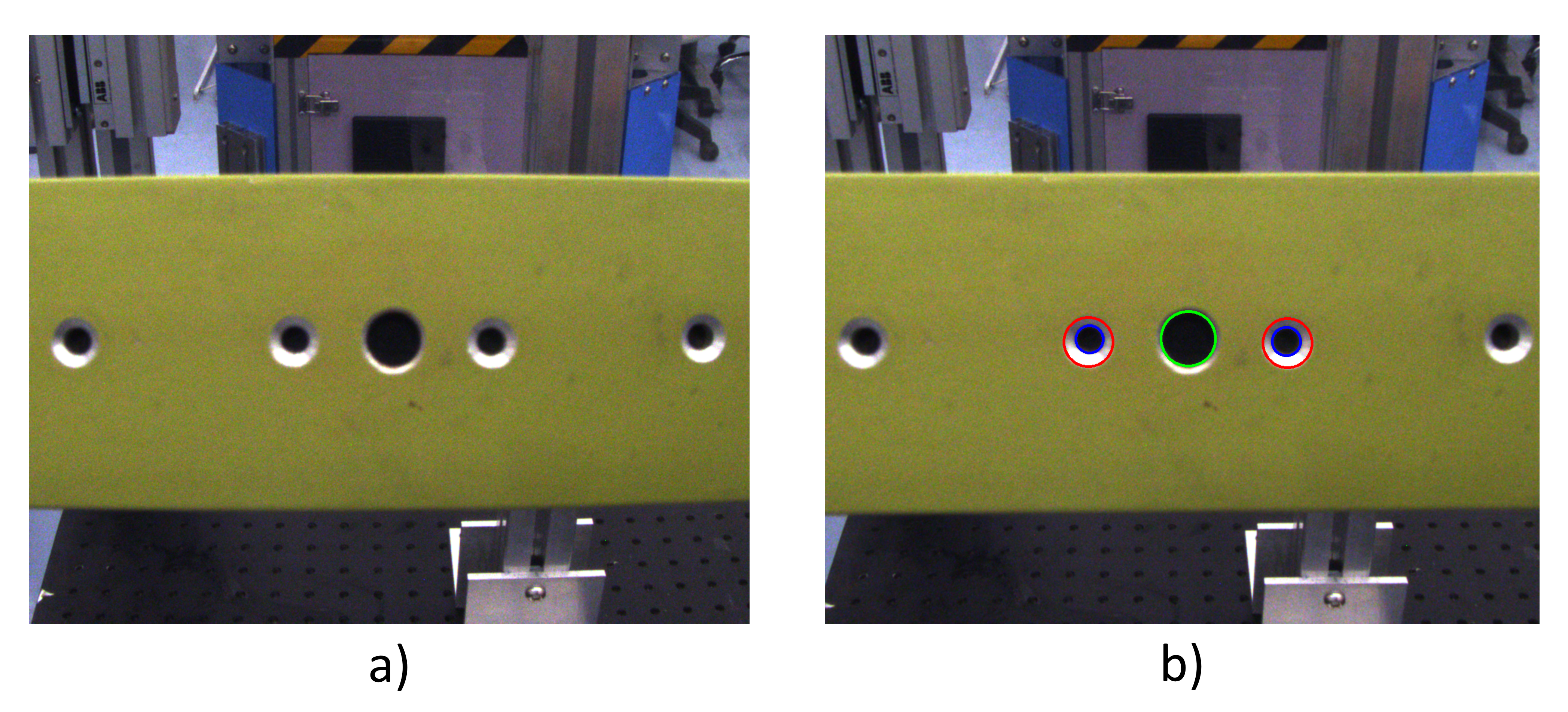}
\caption{ED-Circles applied on a) IDS images. b) Reliable detection of countersinks requires the robot to pause at each hole due to high latency and motion blur.}
\label{fig:ids_inspect}
\end{figure}

In addition to high precision, our work provides the advantage of the swift execution time of the inspection task, with the capability of the NVS sweeping the workpiece at a high speed of 0.5 m/s. Table \ref{speeds} reports the sweep time at different NVS speeds and illustrates that the 60 countersinks workpiece is scanned in nearly 3 seconds. Previous methods, in contrast, require the robot to pause on each countersink to perform the inspection before proceeding to the next hole. We quantify this by inspecting the holes using the IDS frames, and hole detection was carried out using ED-Circles and brute-force matching for outlier removal, in which these experiments mimic the works of Yu et al. \cite{yu2019vision}. This is illustrated in Figure \ref{fig:ids_inspect}. During the inspection, the IDS camera paused on each hole for 0.5 seconds to carry on the inspection algorithm, and the motion between holes takes nearly 0.2 seconds. In total, the inspection task takes almost 42 seconds. One important aspect of our inspection pipeline is the computational time. Table \ref{table:exec_time} reports the total execution time for the proposed inspection pipeline algorithms. As 0.033 seconds are required to inspect each countersink, the total execution time of the neuromorphic vision-based inspection task is 4.98 seconds, reducing the inspection time by nearly a factor of 10. This also shows that the proposed work can be utilized to run online. It is worth mentioning that our algorithm was written in Python, and real-time capabilities can be achieved if written in C++.

\begin{table}[ht]
\centering
\caption{Sweep time at different NVS speeds for a 1.5 m countersinks workpiece.}
\begin{tabular}{|c|c|c|c|c|c|}
\hline
\textbf{Speed (m/s)} & 0.05 & 0.1 & 0.2 & 0.3 & 0.5 \\ \hline
\textbf{Sweep Time (s)} & 30 & 15 & 7.5 & 5 & 3 \\ \hline
\end{tabular}
\label{speeds}
\end{table}

\begin{table}[ht]
\centering
\caption{Execution times for the system blocks.}
\begin{tabular}{|c|c|}
\hline
\textit{\textbf{Algorithm}} & \textit{\textbf{Execution Time (s)}} \\ \hline
\begin{tabular}[c]{@{}c@{}}Image of Warped Events\\ Reconstruction\end{tabular} & 0.0084 \\ \hline
\begin{tabular}[c]{@{}c@{}}Mean Shift\\ Clustering\end{tabular} & 0.0178 \\ \hline
\begin{tabular}[c]{@{}c@{}} Huber-Based\\ Circle Fitting\end{tabular} & 0.007 \\ \hline
 Depth Estimation & 0.000013 \\ \hline
\textbf{Total} & 0.0333 \\ \hline
\end{tabular}
\label{table:exec_time}
\end{table}

\section{Conclusions and Future Work} \label{sec:conc}
In this paper, and for the first time, high-speed neuromorphic vision-based countersink inspection was proposed for automated assembly lines. The proposed approach leverages the neuromorphic vision sensor's unprecedented asynchronous operation to create sharp, deblurred images through motion compensation. Due to the NVS unorthodox nature, a robust circle detection algorithm was devised to detect the countersinks in the IWEs. To prove the feasibility of neuromorphic vision for high-speed inspection, we have tested this paper's work in various experiments for a low-resolution NVS reaching speeds up to 0.5 m/s. The high quality of the IWEs and the proposed circle detector played a major role in determining the precision of the proposed system. The experimental results show that over 60 countersinks can be inspected in less than 5 seconds while maintaining a precision of 0.025 mm. For future work, we aim to develop an event-based fiducial marker detection algorithm to utilize the NVS solely for the whole inspection procedure, starting from the initial alignment of the sensor with the inspected workpiece to its high-speed sweep. On the other hand, while the introduced motion compensation-circle detection frameworks were utilized for inspection, these algorithms will be further deployed for robotic positioning systems and visual servoing in automated manufacturing processes such as precision drilling, deburring, and various other robotic manufacturing opportunities.

\begin{acknowledgements}
This work was supported by by the Advanced Research and Innovation Center (ARIC), which is jointly funded by STRATA Manufacturing PJSC (a Mubadala company), Khalifa University of Science and Technology in part by Khalifa University Center for Autonomous Robotic Systems under Award RC1-2018-KUCARS, and Sundooq Al Watan under Grant SWARD-S22-015.
\end{acknowledgements}

%



\section*{Author Contribution}
\textbf{Mohammed Salah} Conceptualization, Methodology, Software, Investigation, Data collection, Experimentation, Writing. \textbf{Abdulla Ayyad} Conceptualization, Methodology, Software, Investigation, Data collection, Experimentation, Writing. \textbf{Mohammed Ramadan} End-effector Design, Software, Writing. \textbf{Yusra Abdulrahman} Technical Advising, Writing. \textbf{Dewald Swart} Conceptualization, Software. \textbf{Abdelqader Abusafieh} Methodology, Technical Advising. \textbf{Lakmal Seneviratne} Supervision, Funding acquisition, Technical Advising. \textbf{Yahya Zweiri} Project management, Funding acquisition, Review and Editing.

\section*{Conflict of interest}
The authors declare that they have no conflict of interest.


%
%

\bibliographystyle{unsrt}
\bibliography{template.bib}

\end{document}